\documentclass[10pt]{article}
\usepackage[utf8]{inputenc}
\usepackage[T1]{fontenc}
\usepackage{lmodern}
\usepackage[a4paper,margin=0.86in,headheight=14pt]{geometry}
\usepackage{microtype}
\usepackage{graphicx}
\usepackage{booktabs,longtable,tabularx,array,multirow,calc}
\usepackage{amsmath,amssymb,bm}
\usepackage[dvipsnames]{xcolor}
\definecolor{AstronexBlue}{HTML}{143C69}
\definecolor{AstronexCyan}{HTML}{087D96}
\definecolor{AstronexPanel}{HTML}{F2F5FB}
\usepackage{hyperref}
\hypersetup{colorlinks=true,linkcolor=black,citecolor=black,urlcolor=black,pdftitle={Astronex-World 1.0: Real-Time Interactive World Model Foundation},pdfauthor={Xin Zhou and Cong Miao}}
\usepackage{caption}
\usepackage{tcolorbox}

\usepackage{float,placeins}
\usepackage{enumitem}
\setlist{nosep,leftmargin=1.7em}
\usepackage{titlesec}
\titleformat{\section}{\Large\bfseries\color{black}}{\thesection}{0.65em}{}
\titleformat{\subsection}{\large\bfseries\color{black}}{\thesubsection}{0.6em}{}
\titleformat{\subsubsection}{\normalsize\bfseries\color{black}}{\thesubsubsection}{0.5em}{}
\titlespacing*{\section}{0pt}{2.2ex plus .5ex}{1.0ex}
\titlespacing*{\subsection}{0pt}{1.8ex plus .4ex}{0.8ex}
\usepackage{fancyhdr}
\AtBeginEnvironment{longtable}{\small\setlength{\tabcolsep}{3.5pt}}
\providecommand{\tightlist}{\setlength{\itemsep}{0pt}\setlength{\parskip}{0pt}}

\title{\vspace{-1.5em}\textbf{\LARGE Astronex-World 1.0: Real-Time Interactive\\World Model Foundation}}
\author{Xin Zhou\textsuperscript{1}\qquad Cong Miao\textsuperscript{2}\\[3pt]
\small \textsuperscript{1}Astronex Robotics\qquad \textsuperscript{2}Nanjing University of Information Science and Technology\\[-2pt]
}
\date{Technical Report, 2026}

\begin{document}
\maketitle
\sloppy
\emergencystretch=2em
\begin{tcolorbox}[
  colback=AstronexPanel,
  colframe=AstronexPanel,
  boxrule=0pt,
  arc=3mm,
  left=5mm,right=5mm,top=4mm,bottom=4mm]
\small
We present Astronex-World 1.0, an open controllable video world-model foundation. Given a text prompt (text-to-video) or an initial observation (image-to-video), the model predicts future visual states under frame-aligned camera trajectories, continuous actions, and an embodiment identifier, and it accepts text events inserted at a specified position of a rollout. The family provides a bidirectional model for full-context generation and supervision and a causal model with block-causal attention and cross-block KV caching for persistent generation. Both inherit the Wan2.2-TI2V-5B visual prior \cite{ref1}. PRoPE injects camera intrinsics and extrinsics \cite{ref6}, while a 64-dimensional action stream modulates every Transformer layer. A five-stage training path develops bidirectional camera and action control, converts the backbone to block-causal generation, transfers the dense UniPC trajectory to a few-step student, restores mixed-domain dynamics, and completes asymmetric DMD/DMD2 distribution matching. The causal model generates $832\times480$ video at 24 fps with eight-step UniPC, an eight-latent-frame block, local history, and persistent sink frames. All five training stages run on two NVIDIA L20 48 GB GPUs, and the causal model streams in real time on one. It scores 73.5 on WBench Navi and 70.0 on WBench Full. On Full, this 5B model is above the 13.6B LongCat-Video and the 14B Helios, within one point of the 22B LTX-2.3, and above YUME 1.5, which is post-trained from the same 5B Wan2.2 prior on NVIDIA A100 GPUs. The reserved action input and output interfaces allow post-training for embodied intelligence and autonomous driving.

\vspace{0.9em}
\noindent{\sffamily\bfseries Website:}~\href{https://world.astronex.com.cn/}{\textcolor{AstronexCyan}{\texttt{https://world.astronex.com.cn}}}\\
{\sffamily\bfseries GitHub:}~\href{https://github.com/Astronex-Robotics/Astronex-World}{\textcolor{AstronexCyan}{\texttt{https://github.com/Astronex-Robotics/Astronex-World}}}\\
{\sffamily\bfseries Hugging Face:}~\href{https://huggingface.co/Astronex-Lab/Astronex-World}{\textcolor{AstronexCyan}{\texttt{https://huggingface.co/Astronex-Lab/Astronex-World}}}
\end{tcolorbox}

\vspace{0.4em}
\begin{figure}[H]
\centering
\includegraphics[width=\textwidth]{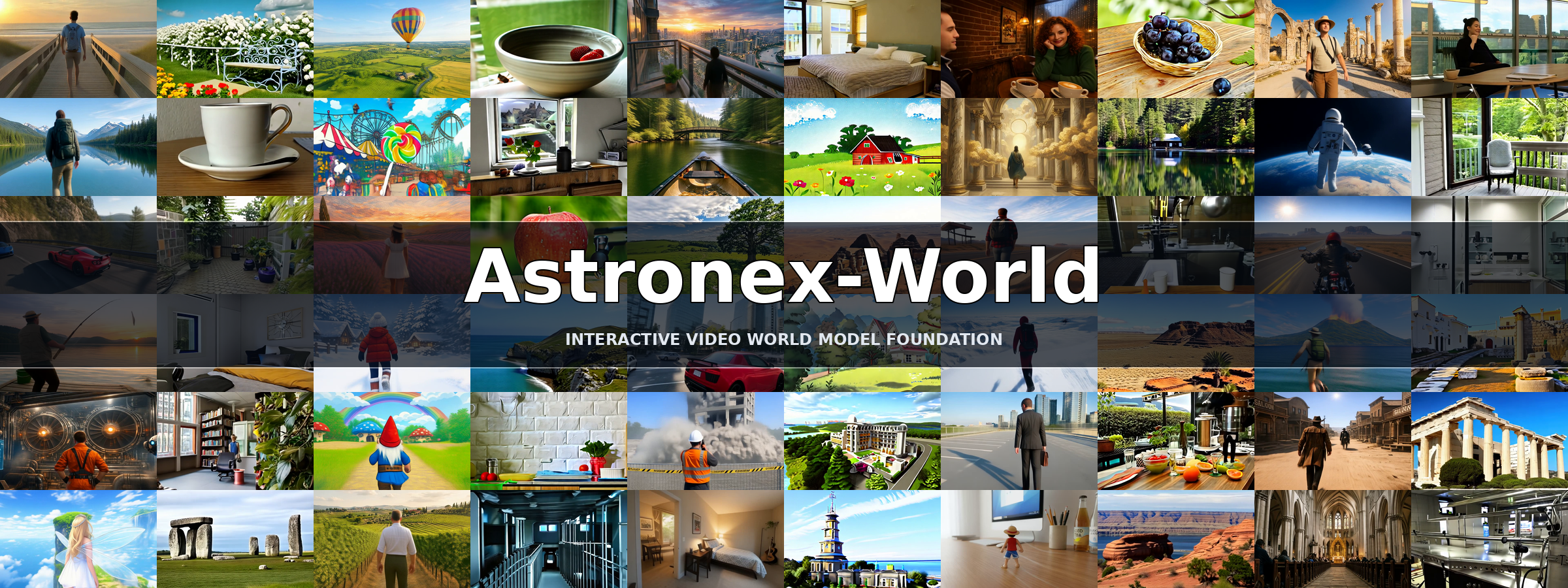}
\caption{Astronex-World 1.0 is a 5B controllable video world-model foundation with bidirectional and causal releases. Frames are sampled from videos generated by the model in interactive, image-to-video, and text-to-video settings.}
\label{fig:teaser}
\end{figure}

\section{Introduction}\label{introduction}

World models learn controllable dynamics between observations, actions, and future observations \cite{ref43}. A video world model predicts how a scene evolves under camera motion, actions, and events, and a shared foundation of this kind can reuse large-scale generative priors instead of being trained separately for every task. This report presents such a foundation. Its action interface is reserved for post-training, through which the foundation can be connected to embodied intelligence and autonomous driving.

Three tensions make this conversion difficult. First, pretrained video diffusion Transformers typically use bidirectional attention: each position may read the entire noisy sequence. Interactive generation is causal and may access only history. Second, text and a first frame cannot fully specify executable control; the model must understand time-aligned camera geometry and continuous actions. Third, autoregressive rollout feeds model errors back into the context, causing motion decay, geometric drift, color shift, and gradual loss of detail.

Astronex-World starts from the Wan2.2-TI2V-5B visual prior \cite{ref1} and builds two complementary model forms around a shared control interface. The bidirectional model retains full temporal context for high-quality conditional generation, control adaptation, and teacher supervision. The causal model converts temporal attention into a block lower-triangular pattern, reuses only historical KV states, and generates continuously with few-step UniPC sampling. Cameras are represented by PRoPE, while actions are represented by a 64-dimensional continuous vector and embodiment ID that modulate every layer. The system directly supports interactive video generation and preserves input, output, and checkpoint interfaces for domain post-training. The 64-dimensional action input and the action-sequence output are kept open so that driving and robot data can be added in post-training (Section~\ref{downstream-adaptation}).

Our main contributions are:

\begin{enumerate}
\def\labelenumi{\arabic{enumi}.}
\tightlist
\item
  A shared 5B controllable video backbone with bidirectional and causal releases that supports both text-to-video and image-to-video generation. The former targets high-quality offline generation and supervision; the latter targets cacheable, persistent, few-step interaction.
\item
  A unified control interface that injects complete camera intrinsics and extrinsics through PRoPE and modulates every visual layer with a 64-dimensional continuous action stream and embodiment ID. An event interface inserts a text event at a specified position of an ongoing rollout, and a post-trainable action-sequence output interface is reserved for embodied intelligence and autonomous driving.
\item
  A five-stage training recipe (Stages I--V): bidirectional control adaptation, block-causal teacher forcing, online UniPC trajectory distillation, mixed-domain causal SFT, and asymmetric DMD. The whole recipe runs on two NVIDIA L20 48 GB GPUs.
\item
  Evaluation on WBench (Navi 158 and Full 289) and partial VBench 1.0 runs, and a comparison with open models of 4B to 22B parameters showing that a 5B model post-trained on two L20 GPUs reaches the WBench Full score range of 13.6B to 22B models.
\end{enumerate}

\section{Related Work}\label{related-work}

\subsection{World Models}\label{world-models}

World models learn a predictive model of an environment in which an agent can act \cite{ref43}. Latent-dynamics agents such as DreamerV3 learn these models for control \cite{ref44}, while generative world models synthesize observations directly. Genie learns playable environments from unlabeled video \cite{ref45}, and GameNGen and DIAMOND simulate games with diffusion models \cite{ref46,ref47}. Matrix-Game 2.0 and Yume generate interactive open-domain worlds \cite{ref48,ref49}, and WorldPlay, LingBot-World, and Genie 3 target real-time interaction with long-horizon consistency \cite{ref34,ref33,ref35}. In autonomous driving, GAIA-1 and GAIA-2 generate controllable driving video \cite{ref50,ref51}, Vista generalizes across driving scenes and control inputs \cite{ref52}, and Cosmos provides world foundation models for physical AI \cite{ref53}. For embodied agents, UniSim simulates real-world interaction from video \cite{ref54}, UniPi derives policies from text-guided video generation \cite{ref55}, and V-JEPA 2 uses a self-supervised video model for prediction and planning \cite{ref56}. Astronex-World follows the generative line with one camera- and action-conditioned video backbone that serves as a general foundation.

\subsection{Video Diffusion and Video DiTs}\label{video-diffusion-and-video-dits}

Diffusion models learn high-dimensional distributions through iterative denoising \cite{ref2}. DiT replaces the conventional denoising backbone with a Transformer operating on latent patches and scales effectively with model capacity \cite{ref3}. Wan provides an open video foundation family with a video VAE, DiT backbone, and multiple generation tasks \cite{ref1}. Astronex-World retains the Wan2.2-TI2V-5B latent representation and 30-layer DiT backbone rather than repeating large-scale video pretraining, and introduces world-model controls through low-rank adaptation \cite{ref5} and newly initialized modules. The objective follows a flow-based parameterization; the general formulation is described by Flow Matching and rectified flow \cite{ref4,ref39}, and the timestep shift follows Stable Diffusion 3 \cite{ref38}.

\subsection{Causal Video Generation and Long-Term Memory}\label{causal-video-generation-and-long-term-memory}

In a Transformer decoder, masked self-attention blocks access to future positions and enables autoregressive computation \cite{ref28}, whereas unmasked temporal attention aggregates context in both directions. Full spatiotemporal attention benefits global video quality \cite{ref1}, but a streaming world model must rely only on history. CausVid demonstrates that a bidirectional video diffusion model can supervise a causal few-step generator with KV caching \cite{ref7}. Self Forcing conditions training on self-generated history to reduce train-test mismatch \cite{ref8}, while Diffusion Forcing assigns different noise levels to sequence tokens and unifies causal prediction with diffusion \cite{ref22}. MAGI-1 and SkyReels-V2 scale autoregressive and diffusion-forcing video generation to long outputs \cite{ref59,ref60}.

Causal Forcing studies the pointwise mapping required when transferring a bidirectional teacher to an autoregressive student and uses teacher-forced AR diffusion and causal-ODE initialization \cite{ref20}. Causal Forcing++ replaces offline causal-ODE data with online causal consistency distillation \cite{ref21,ref23}. In Astronex-World, CF/CF++ covers only causal initialization and online trajectory transfer (Stages II and III in Section~\ref{training}). Mixed-data SFT and DMD/DMD2 are separate post-training stages (Stages IV and V) and are not counted as additional CF++ stages.

Attention sinks were introduced for streaming Transformers: retaining the KV states of the first tokens mitigates degradation when the context window slides \cite{ref27}. Rolling Forcing combines a progressive-noise window, long rollout, and attention sinks to reduce autoregressive accumulation \cite{ref9}. LongLive similarly balances long-range consistency and efficiency with short windows and frame sinks \cite{ref10}. Other work keeps long-range scene memory explicitly: Context-as-Memory retrieves past frames by camera field-of-view overlap \cite{ref40}, WorldMem stores past frames and their states in a memory bank \cite{ref41}, and VMem indexes past views with surfels \cite{ref42}. Astronex-World uses block-causal attention, a local KV window, and a fixed initial sink. Random rollout lengths and rolling-forcing probability expose training gradients to different history positions.

\subsection{Geometric and Action Control}\label{geometric-and-action-control}

Camera-controlled video generation often encodes the camera trajectory as an additional condition: CameraCtrl uses Pl\"ucker embeddings \cite{ref57}, and MotionCtrl conditions on camera poses and object trajectories \cite{ref58}. PRoPE instead combines camera intrinsics and extrinsics into a projective positional representation so that attention operates in relative camera geometry \cite{ref6}. Astronex-World inserts this mechanism into all 30 Transformer layers and explicitly reconciles c2w, w2c, and normalized-intrinsics conventions in its WBench adapter. The action path accepts 64-dimensional continuous conditions and 32 embodiment IDs.

\subsection{Few-Step Sampling and Distribution Matching}\label{few-step-sampling-and-distribution-matching}

Distribution Matching Distillation (DMD) approximates a distribution-level KL gradient using the score difference between real and generated distributions \cite{ref24}. DMD2 improves few-step training through separate generator and fake-score optimization and multistep training \cite{ref11}; related work extends distribution matching to video \cite{ref12}. Astronex-World uses the bidirectional control model (restore-2200) as a real-score teacher and adds a motion-preservation term on mixed physical data. Inference uses UniPC \cite{ref13}. Distillation targets four-step generation; the released model can generate with four steps, and eight steps give the best quality.

\subsection{Training Infrastructure}\label{training-infrastructure}

We use minWM as the engineering framework for data processing, distributed training, distillation, checkpoint merging, and streaming inference \cite{ref19}. Astronex-World integrates Wan2.2-TI2V-5B and adds its own bidirectional/causal checkpoint lineage, PRoPE camera control, 64-dimensional action modulation, embodiment IDs, reserved action output, long-window/sink behavior, and WBench/VBench adapters. minWM is the implementation framework; Astronex-World is the trained, released, and evaluated model family.

A capability-level comparison of the released interactive world models discussed above, covering model form, control interface, streaming behavior and output resolution, appears in Section~\ref{capability-comparison}.

\section{System Overview}\label{system-overview}

\begin{figure}[H]
\centering
\includegraphics[width=\linewidth]{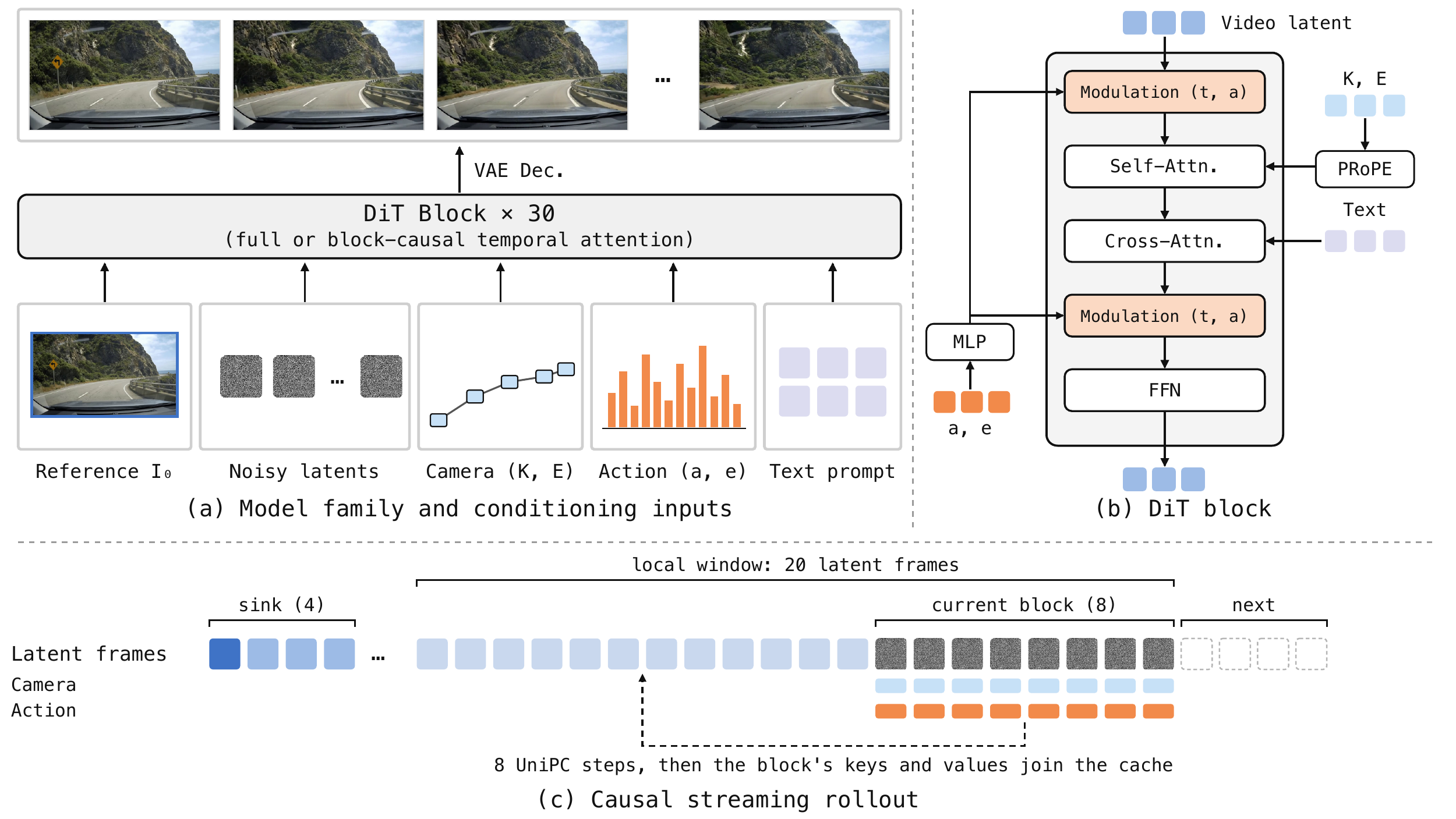}
\caption{Astronex-World model and inference pipeline. (a) One 30-block video DiT, used with full temporal attention (bidirectional checkpoint) or block-causal attention (causal checkpoint), takes a reference frame, noisy latents, per-frame camera and action conditions, and a text prompt, and decodes the result with the VAE. (b) Inside each block, camera matrices enter self-attention through PRoPE, the text enters through cross-attention, and the action vector with its embodiment ID is mapped by an MLP into the scale, shift, and gate modulation that also carries the timestep. (c) The causal model generates eight latent frames per block with eight UniPC steps, attends to four sink frames and a 20-frame local window, and adds each finished block to the KV cache. Frames in (a) are from an action-controlled driving rollout on the project website.}
\label{fig:project-2}
\end{figure}

The system has four layers. The data layer aligns video, text, first-frame input, camera matrices, actions, and embodiment IDs to latent-frame tensors. The model layer wraps the Wan2.2 bidirectional or causal trunk together with PRoPE, action modulation, and an optional output head. The training layer selects bidirectional diffusion, AR diffusion, causal consistency distillation, SFT, or score distillation, corresponding to Stages I--V in Section~\ref{training}. The inference and evaluation layer handles checkpoint overlays, KV caches, history windows, samplers, and benchmark protocols. All stages share the same 20-frame latent geometry and control interface.

{\def\LTcaptype{table}
\begin{longtable}[]{@{}
  >{\raggedright\arraybackslash}p{(\linewidth - 2\tabcolsep) * \real{0.5000}}
  >{\raggedright\arraybackslash}p{(\linewidth - 2\tabcolsep) * \real{0.5000}}@{}}
\caption{Configuration of the released Astronex-World-5B model.}\label{tab:model-config}\\
\toprule\noalign{}
\begin{minipage}[b]{\linewidth}\raggedright
Item
\end{minipage} & \begin{minipage}[b]{\linewidth}\raggedright
Configuration
\end{minipage} \\
\midrule\noalign{}
\endfirsthead
\toprule\noalign{}
\begin{minipage}[b]{\linewidth}\raggedright
Item
\end{minipage} & \begin{minipage}[b]{\linewidth}\raggedright
Configuration
\end{minipage} \\
\midrule\noalign{}
\endhead
\bottomrule\noalign{}
\endlastfoot
Model & Astronex-World-5B \\
Forms & Bidirectional control model and causal few-step model \\
Base model & Wan2.2-TI2V-5B \\
Parameters & 5,351,000,000 \\
Weight precision & bfloat16 \\
DiT depth & 30 layers \\
Hidden width / heads & 3072 / 24 \\
FFN width & 14,336 \\
Latent channels & 48 \\
VAE compression & $4\times$ temporal, $16\times16$ spatial \\
Native conditions & Text, first frame, camera intrinsics/extrinsics, 64-D actions \\
Camera injection & PRoPE in all 30 layers \\
Action space & 64-D continuous action, 32 embodiment IDs \\
Action output & Open post-training head, 64-D per latent frame; enabled when training action-sequence decoding \\
Causal context & 20 latent frames plus four persistent sink frames \\
Default sampler & UniPC, eight steps, CFG 3.0 \\
Inference block & Eight latent frames \\
Output & $832\times480$ and $1280\times704$, 24 fps (benchmarks use $832\times480$) \\
\end{longtable}
}

The bidirectional checkpoint family centers on restore-2200 and uses full temporal attention for offline generation, domain adaptation, and DMD score supervision. The causal family continues from the shared controls through AR conversion, few-step distillation, and mixed-data post-training. WBench and VBench evaluate the causal model; inference does not split one generation across the bidirectional teacher and causal student.

\section{Data}\label{data}

\subsection{Camera-Control Data}\label{camera-control-data}

Stage I camera adaptation and camera recovery, as well as the causal Stages II and III, use the Control2V camera-control data released with minWM \cite{ref19,ref30}. Two direction-augmented indices are built from the same data: \texttt{Control2V\_\allowbreak{}backdown} (Stage I-a) and \texttt{Control2V\_\allowbreak{}combined} (Stages I-d and II). They are re-indexed views of Control2V, not separate datasets. Each training sample provides video latents, camera intrinsics, and framewise extrinsics. RGB clips are encoded by the Wan2.2 VAE into 48-channel video latents at the model's native training resolution.

\subsection{Mixed Physical and Action Data}\label{mixed-physical-and-action-data}

The bidirectional action branch (Stage I-b) uses DROID \cite{ref26} to learn the 64-dimensional action condition and embodiment identifier. Mixed-domain adaptation (Stages IV and V) combines Control2V, CrossFPS, DrivingDojo, and NVIDIA PhysicalAI. Table~\ref{tab:datasets} lists the datasets, what they supervise, and the stages that use them.

{\def\LTcaptype{table}
\begin{longtable}[]{@{}
  >{\raggedright\arraybackslash}p{(\linewidth - 6\tabcolsep) * \real{0.24}}
  >{\raggedright\arraybackslash}p{(\linewidth - 6\tabcolsep) * \real{0.27}}
  >{\raggedright\arraybackslash}p{(\linewidth - 6\tabcolsep) * \real{0.33}}
  >{\raggedright\arraybackslash}p{(\linewidth - 6\tabcolsep) * \real{0.16}}@{}}
\caption{Training datasets, their supervision, and the stages that use them.}\label{tab:datasets}\\
\toprule\noalign{}
\begin{minipage}[b]{\linewidth}\raggedright
Dataset
\end{minipage} & \begin{minipage}[b]{\linewidth}\raggedright
Modality and supervision
\end{minipage} & \begin{minipage}[b]{\linewidth}\raggedright
Primary role
\end{minipage} & \begin{minipage}[b]{\linewidth}\raggedright
Stages
\end{minipage} \\
\midrule\noalign{}
\endfirsthead
\toprule\noalign{}
\begin{minipage}[b]{\linewidth}\raggedright
Dataset
\end{minipage} & \begin{minipage}[b]{\linewidth}\raggedright
Modality and supervision
\end{minipage} & \begin{minipage}[b]{\linewidth}\raggedright
Primary role
\end{minipage} & \begin{minipage}[b]{\linewidth}\raggedright
Stages
\end{minipage} \\
\midrule\noalign{}
\endhead
\bottomrule\noalign{}
\endlastfoot
Control2V / minWM-data \cite{ref19,ref30} & Video, camera intrinsics, camera extrinsics & Camera trajectories and projective geometry & I-a, I-c, I-d, II, III, IV, V \\
DROID \cite{ref26} & Real robot trajectories and actions & Bidirectional action adaptation and embodiment conditioning & I-b \\
CrossFPS \cite{ref14} & First-person video with aligned controls & Action-conditioned navigation and game motion & IV, V \\
DrivingDojo \cite{ref15} & Driving video and road scenarios & Ego motion, traffic events, and road dynamics & IV, V \\
NVIDIA PhysicalAI \cite{ref16} & Synthetic embodied robot scenes & Contact, manipulation, and embodied motion & IV, V \\
\end{longtable}
}

\section{Method}\label{method}

\subsection{Text-to-Video and Image-to-Video Generation}\label{t2v-i2v-generation}

\begin{figure}[H]
\centering
\includegraphics[width=\linewidth]{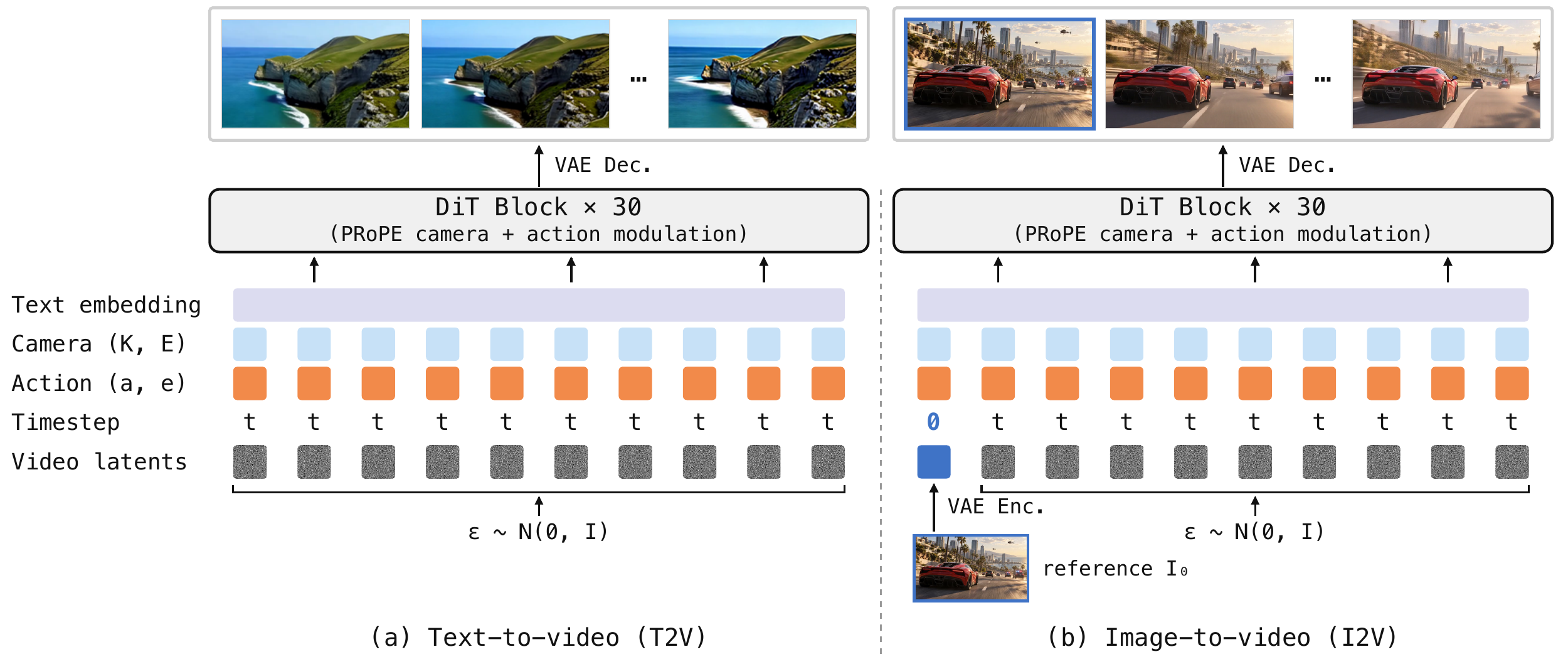}
\caption{T2V and I2V with the same Astronex-World DiT. Each column is one latent frame. The prompt is a global condition read through cross-attention; camera \((K,E)\) and action \((a,e)\) are aligned to latent frames and enter every block through PRoPE and action modulation. (a) T2V: every latent frame starts from Gaussian noise and shares the time \(t\). (b) I2V: the reference frame \(I_0\) is VAE-encoded into the first latent frame, which stays clean at time 0 and is excluded from the loss; only the remaining frames are denoised. The DiT is called at every UniPC step, and the VAE decoder maps the final latent to RGB. Frames in (a) come from a VBench T2V video and frames in (b) from an I2V clip on the project website.}
\label{fig:t2v-i2v-pipeline}
\end{figure}

Astronex-World inherits the unified text-image-to-video design of Wan2.2-TI2V-5B \cite{ref1}. Text-to-video (T2V) and image-to-video (I2V) use the same weights; they differ only in how the latent sequence is initialized and which latent frames are denoised (Figure~\ref{fig:t2v-i2v-pipeline}).

\paragraph{Latent video and conditions.} An RGB clip with \(N=4(F-1)+1\) frames is encoded by the Wan2.2 video VAE \(\mathcal{E}\) into a latent \(z\in\mathbb{R}^{F\times 48\times H/16\times W/16}\). The first RGB frame is encoded alone into the first latent frame, and every later latent frame covers four RGB frames. A \(1\times2\times2\) patch embedding turns each latent frame into spatial tokens, and the decoder \(\mathcal{D}\) maps latents back to RGB. The prompt is encoded once by the frozen text encoder into \(c_{\mathrm{text}}\), which every block reads through cross-attention. The control condition \(c_{\mathrm{ctrl}}=\{K_i,E_i,a_i\}_{i=1}^{F}\) and the embodiment ID \(e\) are aligned to latent frames: camera intrinsics \(K_i\) and extrinsics \(E_i\) enter self-attention through PRoPE, and the 64-D action \(a_i\) enters through per-frame modulation (Section~\ref{camera-action-and-event-conditioning}).

\paragraph{Flow parameterization.} For a clean latent \(z_0\), Gaussian noise \(\epsilon\sim\mathcal{N}(0,I)\), and a time index \(t\) with shifted noise level \(\sigma_t\in[0,1]\) \cite{ref38}, the noisy latent and the training objective are
\begin{equation}
z_t=(1-\sigma_t)\,z_0+\sigma_t\,\epsilon,\qquad
\mathcal{L}_{\mathrm{flow}}=\mathbb{E}\,\bigl\|v_\theta(z_t,t,c_{\mathrm{text}},c_{\mathrm{ctrl}})-(\epsilon-z_0)\bigr\|_2^2 .
\label{eq:flow}
\end{equation}
At inference, UniPC \cite{ref13} integrates the predicted velocity from \(\sigma=1\) to \(\sigma=0\). Each solver step calls the DiT once per guidance branch, and the classifier-free guided velocity \cite{ref36} is \(\hat v=v_\theta^{\varnothing}+w\,(v_\theta^{c}-v_\theta^{\varnothing})\), where \(v_\theta^{\varnothing}\) is conditioned on a fixed negative prompt and \(w=3\) by default. After the last step, the frames are \(\hat X=\mathcal{D}(z_0)\).

\paragraph{Text-to-video.} T2V has no visual reference. All \(F\) latent frames start from noise, \(z_{t}=\epsilon\) at \(\sigma_t=1\), and in the bidirectional model all frames share the same time \(t\) at every solver step. The prompt fixes the scene, objects, and style; when camera or action trajectories are given, they decide how the viewpoint and the controlled agent change from frame to frame. VBench T2V videos are generated this way.

\paragraph{Image-to-video.} I2V adds a reference observation \(I_0\), which becomes the first frame of the output. It is encoded as \(z_{\mathrm{ref}}=\mathcal{E}(I_0)\) and placed in the first latent frame. With the frame mask \(m\in\{0,1\}^F\), where \(m_1=1\) and \(m_i=0\) for \(i>1\), the model input and the per-frame times are
\begin{equation}
z_t=m\odot z_{\mathrm{ref}}+(1-m)\odot\bigl((1-\sigma_t)\,z_0+\sigma_t\,\epsilon\bigr),\qquad
t_i=(1-m_i)\,t .
\label{eq:i2v}
\end{equation}
The reference frame therefore enters every block at time 0 as clean context. Future frames attend to it, but it is never updated: after each UniPC step the first latent frame is reset to \(z_{\mathrm{ref}}\). Training applies the flow loss only to the generated frames,
\begin{equation}
\mathcal{L}_{\mathrm{I2V}}=\mathbb{E}\,\bigl\|(1-m)\odot\bigl(v_\theta(z_t,\mathbf{t},c_{\mathrm{text}},c_{\mathrm{ctrl}})-(\epsilon-z_0)\bigr)\bigr\|_2^2 ,
\label{eq:i2v-loss}
\end{equation}
where \(\mathbf{t}=(t_1,\dots,t_F)\). Appearance, identity, and layout come from \(I_0\), while text, camera, and action determine what happens next. Because both modes share one network, Stage IV keeps the first latent frame clean with probability 0.7 and noises it otherwise, so the same weights are trained for I2V and T2V (Section~\ref{stage-iv}). WBench cases and VBench I2V videos start from this path.

\paragraph{Causal rollout.} The causal checkpoint applies the same rules block by block. In I2V, the first block holds the clean reference latent together with noisy future frames; in T2V, it contains only noise. Each later block starts from fresh noise, is denoised with eight UniPC steps while attending to the cached keys and values of earlier blocks, and is then added to the cache. The reference frame lies inside the four-frame attention sink (Section~\ref{long-horizon-context}), so it stays visible for the whole rollout, and a single I2V start can be extended into an interactive session instead of restarting from an image at every turn.

\subsection{Block-Causal Video Generation}\label{block-causal-video-generation}

The bidirectional model uses unmasked spatiotemporal self-attention: a noisy latent position may attend to both past and future positions in a complete clip. This supports full-context quality, motion, and control learning \cite{ref1}. The bidirectional checkpoint is a released model form, not merely a temporary training teacher.

The causal model replaces temporal attention with a block lower-triangular mask. Positions within the current latent block interact bidirectionally, but cross-block attention is restricted to generated history. Once a block is completed, its keys and values are cached and reused without recomputing the entire past \cite{ref7,ref28}. Training uses four latent frames per block; inference uses eight to reduce boundaries and improve throughput. Generation operates in the compressed video latent space, and the VAE is used only for input encoding and final decoding. The conversion from the bidirectional to the causal model is carried out in Stages II--V (Section~\ref{training}).

\subsection{Camera and Action Conditioning}\label{camera-action-and-event-conditioning}

For each latent frame \(i\), the camera branch receives an intrinsic matrix \(K_i\) and world-to-camera transform \(E_i\). PRoPE constructs a frustum-related positional transformation from the projection matrices and applies it to the attention query, key, value, and output projections \cite{ref6}. In WBench, the provided \texttt{extrinsic} follows a camera-to-world convention whereas training uses world-to-camera, so the adapter inverts every matrix. Pixel-space intrinsics are normalized by the source width and height.

The action branch receives a 64-dimensional continuous vector and embodiment ID per latent frame. A two-layer MLP transforms the action; the embodiment embedding is added and projected into six modulation groups per frame: shift, scale, and residual gate for attention and FFN. These increments are added to the Wan time modulation, while PRoPE independently changes attention geometry. The final action projection is zero-initialized so that insertion initially preserves the base model behavior.

\texttt{action\_\allowbreak{}output} is an open post-training capability of the released framework. Spatial visual tokens are pooled per latent frame and passed through a LayerNorm-MLP decoder to produce an action sequence of shape \texttt{{[}B,F,64{]}}. Downstream post-training can use it as an inverse-dynamics or action-sequence decoder \cite{ref25}. Input actions may be masked and MSE applied only to masked samples to avoid trivial copying. The WBench video-generation checkpoint uses \texttt{use\_\allowbreak{}action=true}, \texttt{action\_\allowbreak{}dim=64}, and \texttt{num\_\allowbreak{}embodiments=32}; action-sequence output is activated during downstream post-training rather than during video-only benchmark inference.

WBench keyboard commands are mapped to CrossFPS-style stick channels and expanded over their interaction intervals. Events and subject actions that cannot be represented by this action space use neutral actions and the event prompts of Section~\ref{event-prompts}. The current adapter exposes one earliest text-switch point; multiple timed events in one case are merged. This interface limitation helps explain the lower event and perspective-switch scores on Full 289.

\subsection{Event Prompts}\label{event-prompts}

\begin{figure}[H]
\centering
\includegraphics[width=\linewidth]{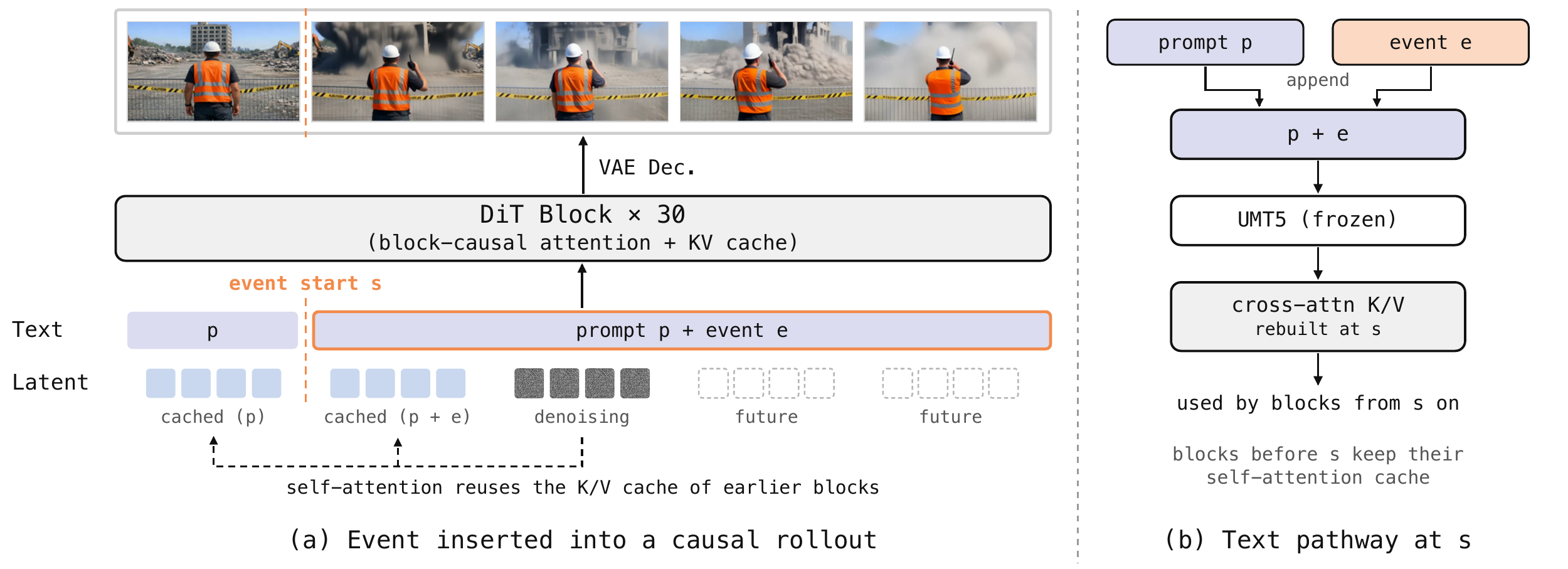}
\caption{Event prompts in the causal model. (a) An event \(e\) is inserted at latent frame \(s\). Blocks before \(s\) were generated with the prompt \(p\) and stay in the KV cache; blocks from \(s\) on are denoised with \(p\oplus e\) while still attending to the cached history, so the event unfolds inside the existing scene. (b) At \(s\), the event text is appended to the prompt, re-encoded by the frozen UMT5 encoder, and the cross-attention keys and values are rebuilt; the self-attention cache of earlier blocks is reused. Frames are from the WBench event-edit Case 265; the position of \(s\) is illustrative.}
\label{fig:event-pipeline}
\end{figure}

Camera and action streams describe how the viewpoint and the agent move; events describe what happens in the scene. The causal model supports events as timed text: an event prompt \(e\) is inserted at a chosen latent frame \(s\) of an ongoing rollout (Figure~\ref{fig:event-pipeline}).

\paragraph{Text condition over time.} Let \(p\) be the prompt that starts the rollout and \(\mathcal{T}\) the frozen UMT5 text encoder. The text condition of latent frame \(i\) is
\begin{equation}
c_{\mathrm{text}}(i)=
\begin{cases}
\mathcal{T}(p), & i<s,\\
\mathcal{T}(p\oplus e), & i\ge s,
\end{cases}
\label{eq:event}
\end{equation}
where \(\oplus\) appends the event text to the prompt. Camera and action conditions are unchanged, so an event can be combined with any trajectory or action stream.

\paragraph{Why the edit stays local.} Three properties of block-causal generation confine the event to the part of the video after \(s\).
\begin{enumerate}
\def\labelenumi{\arabic{enumi}.}
\tightlist
\item
  Frames before \(s\) are finished before the switch. Their keys and values are already in the KV cache and are not recomputed, so the scene established before the event does not change.
\item
  At \(s\) the text is re-encoded and the cross-attention keys and values are rebuilt, so every block from \(s\) on reads \(p\oplus e\). These blocks still attend to the cached history, which ties the event to the existing layout, subjects, and lighting.
\item
  The event is appended rather than substituted. The original prompt keeps describing the scene, and the added text describes only the change. Replacing the prompt would instead ask the model for a different scene.
\end{enumerate}
The bidirectional model denoises the whole clip at once, so it has no finished part before \(s\) and a text change would affect every frame. Event prompts are therefore a feature of the causal form. In an example rollout compared with an otherwise identical event-free run, frames before the switch differ by a mean of 1.8/255 in RGB and frames after it by 9.5/255.

\paragraph{Use for event-driven rollout.} Because the prefix before \(s\) does not depend on the event, the same history can be continued with different events or with no event, and the resulting futures can be compared. For example, a sudden obstacle, a change in weather, or an object interaction can be inserted at a chosen time. Events work best when written in plain language that describes the visible change. The released weights handle events through the text pathway, and the framework additionally provides a dedicated event branch for post-training on data with event annotations.

\subsection{Long-Horizon Context}\label{long-horizon-context}

The causal model attends to a local window of the 20 most recent latent frames, which holds recent history and the current block. In addition, the earliest four latent frames form a persistent attention sink outside this window: they stay in the KV cache and are never evicted as the window slides. A camera-frustum-overlap retrieval option, related to the field-of-view retrieval of Context-as-Memory \cite{ref40}, can substitute high-overlap historical frames when the camera revisits a region. Because discontinuous retrieval can disrupt ordinary forward trajectories, the final WBench configuration enables it only for revisit patterns such as round trips, loops, repeats, L-shapes, and zigzags. Retrieval is disabled for ordinary one-way trajectories and for VBench T2V.

\section{Training}\label{training}

\subsection{Overview of the Five Stages}\label{training-overview}

\begin{figure}[H]
\centering
\includegraphics[width=\linewidth]{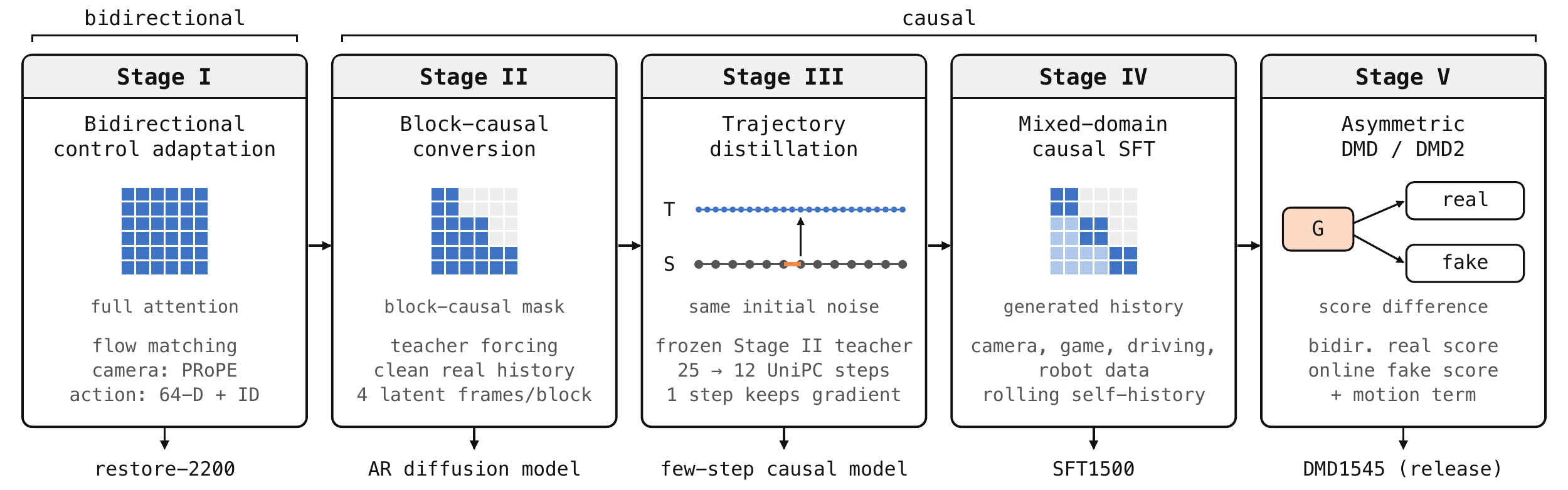}
\caption{The five training stages. Stage I adapts the bidirectional model to camera and action control under full attention. Stage II switches to a block-causal mask and trains with teacher forcing on clean history. Stage III distills a frozen 25-step teacher (T) into a 12-step student (S) that starts from the same noise; one student transition receives gradients. Stage IV fine-tunes on mixed-domain data with generated history (light cells). Stage V matches the generator (G) to a frozen bidirectional real score and an online fake score. The checkpoint produced by each stage is listed below it.}
\label{fig:project-3}
\end{figure}

\begin{figure}[H]
\centering
\includegraphics[width=\linewidth]{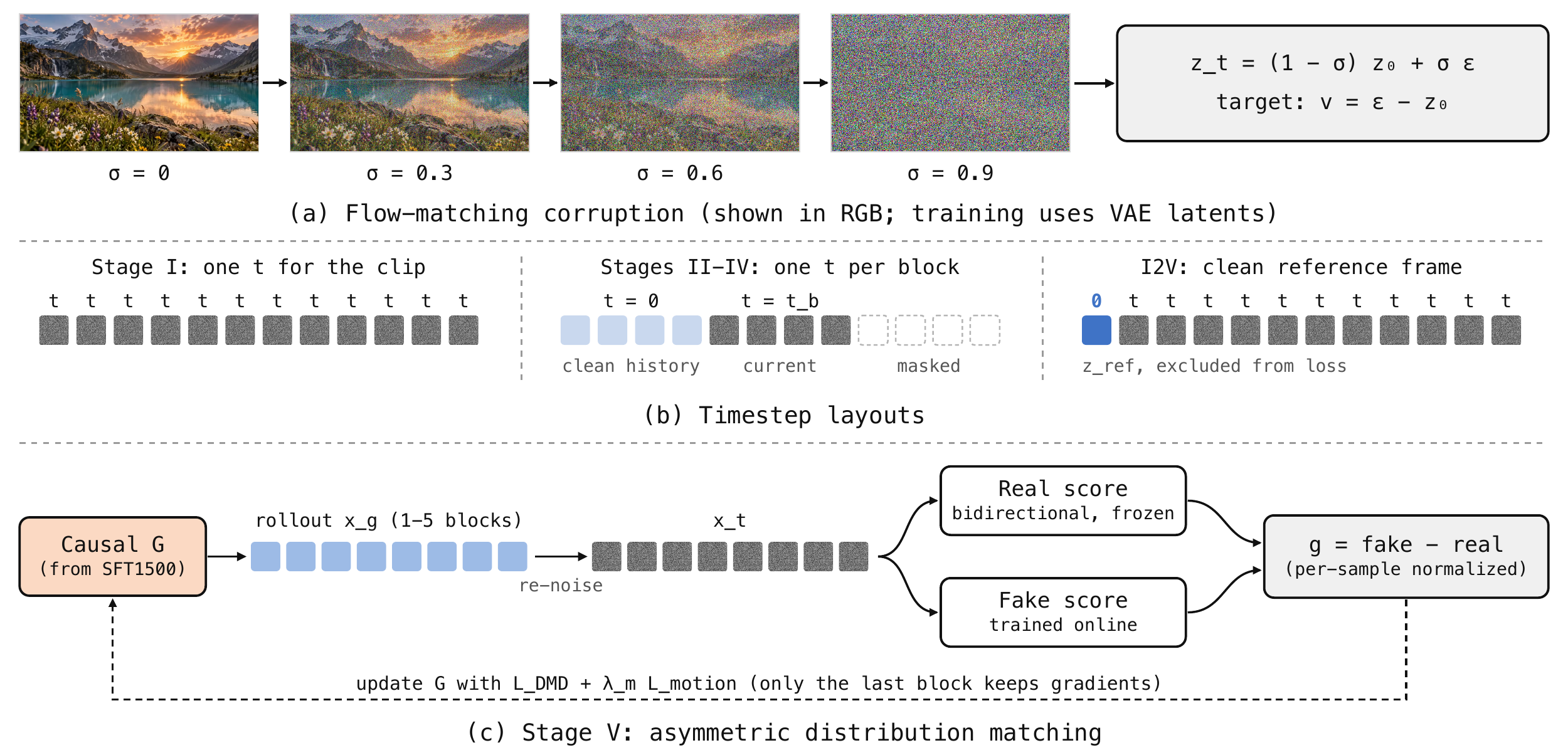}
\caption{Noise layouts and training objectives. (a) Flow-matching corruption, shown in RGB for readability; optimization operates on VAE latents. (b) Stage I shares one timestep across the clip; the causal stages give each four-frame block its own timestep, keep earlier blocks clean, and mask later blocks; in I2V the reference latent is fixed at time 0 and excluded from the loss. (c) In Stage V, a causal rollout is re-noised and scored by the frozen bidirectional real-score model and the online fake-score model; their difference, together with the motion term, updates the generator.}
\label{fig:project-4}
\end{figure}

Astronex-World is trained in five stages, summarized in Table~\ref{tab:stages}. Stage I adapts the bidirectional Wan2.2-TI2V-5B model to camera and action control and produces the bidirectional release. Stages II and III turn this model into a few-step causal generator with the Causal Forcing and Causal Forcing++ components of minWM \cite{ref19,ref20,ref21}: Stage II changes the attention pattern through teacher forcing, and Stage III distills a dense UniPC trajectory into the few-step path used at inference. Stage IV restores subject motion and mixed-domain control with causal SFT, and Stage V applies asymmetric DMD/DMD2 distribution matching \cite{ref11,ref24}, with the Stage I model as the real-score teacher. Figure~\ref{fig:project-3} illustrates the stages, and Figure~\ref{fig:project-4} shows how noise and timesteps are arranged in each objective.

{\def\LTcaptype{table}
\begin{longtable}[]{@{}
  >{\raggedright\arraybackslash}p{(\linewidth - 8\tabcolsep) * \real{0.17}}
  >{\raggedright\arraybackslash}p{(\linewidth - 8\tabcolsep) * \real{0.25}}
  >{\raggedright\arraybackslash}p{(\linewidth - 8\tabcolsep) * \real{0.22}}
  >{\raggedright\arraybackslash}p{(\linewidth - 8\tabcolsep) * \real{0.20}}
  >{\raggedright\arraybackslash}p{(\linewidth - 8\tabcolsep) * \real{0.16}}@{}}
\caption{The five training stages of Astronex-World.}\label{tab:stages}\\
\toprule\noalign{}
Stage & Goal & Attention and objective & Initialization \(\rightarrow\) output & Data \\
\midrule\noalign{}
\endfirsthead
\toprule\noalign{}
Stage & Goal & Attention and objective & Initialization \(\rightarrow\) output & Data \\
\midrule\noalign{}
\endhead
\bottomrule\noalign{}
\endlastfoot
I. Bidirectional control adaptation & Add PRoPE camera control and 64-D action control to the video prior & Full temporal attention; flow matching & Wan2.2-TI2V-5B \(\rightarrow\) restore-2200 (bidirectional release) & Control2V, DROID \\
II. Block-causal conversion & Replace full temporal attention with history-only block attention & Block-causal mask; teacher forcing on clean history & Stage I \(\rightarrow\) multistep AR diffusion model & Control2V \\
III. Online trajectory distillation & Reduce sampling steps under causal attention & 25-step UniPC teacher \(\rightarrow\) 12-step student & Stage II \(\rightarrow\) few-step causal initialization & Control2V \\
IV. Mixed-domain causal SFT & Recover subject motion; learn game, driving, and robot dynamics & Block-causal; window 20, sink 4; rolling forcing & Stage III \(\rightarrow\) SFT1500 & Control2V, CrossFPS, DrivingDojo, PhysicalAI \\
V. Asymmetric DMD/DMD2 & Match the few-step output distribution while keeping motion & Causal self-rollout; bidirectional real score, online fake score & SFT1500 \(\rightarrow\) DMD1545 (causal release) & Same mixture as Stage IV \\
\end{longtable}
}

\subsection{Common Setup}\label{reproducible-retained-lineage}

All stages use \textbf{two NVIDIA L20 48 GB GPUs}, bfloat16 mixed precision, FSDP, gradient checkpointing, 1,000 training time indices, and internal timestep shift 5.0. Training tensors have shape \texttt{{[}1,20,48,30,52{]}}, and causal training blocks contain four latent frames. All LoRA stages use rank 64 and alpha 128. Effective batch size is per-GPU batch times gradient accumulation times the two-GPU data-parallel world size. Table~\ref{tab:stage-config} summarizes the data, initialization, and objective of every step.

{\def\LTcaptype{table}
\begin{longtable}[]{@{}
  >{\raggedright\arraybackslash}p{(\linewidth - 6\tabcolsep) * \real{0.2000}}
  >{\raggedright\arraybackslash}p{(\linewidth - 6\tabcolsep) * \real{0.2800}}
  >{\raggedright\arraybackslash}p{(\linewidth - 6\tabcolsep) * \real{0.2200}}
  >{\raggedright\arraybackslash}p{(\linewidth - 6\tabcolsep) * \real{0.3000}}@{}}
\caption{Per-stage training setup.}\label{tab:stage-config}\\
\toprule\noalign{}
Stage & Data & Initialization & Objective \\
\midrule\noalign{}
\endfirsthead
\toprule\noalign{}
Stage & Data & Initialization & Objective \\
\midrule\noalign{}
\endhead
\bottomrule\noalign{}
\endlastfoot
I-a Camera & Control2V & Wan2.2-TI2V-5B & Flow matching with PRoPE camera control \\
I-b Action & DROID & I-a checkpoint & Flow matching with action and embodiment conditioning \\
I-c Camera recovery & Control2V & I-b checkpoint & Flow matching to restore the camera response \\
I-d Direction enhancement & Control2V & I-c checkpoint & Flow matching for direction control \\
II Teacher forcing & Control2V & Stage I checkpoint & Block-causal teacher forcing \\
III Trajectory distillation & Control2V & Stage II checkpoint & Online UniPC trajectory distillation \\
IV Mixed SFT & Control2V, CrossFPS, DrivingDojo, PhysicalAI & Stage III checkpoint & Causal SFT with rolling forcing \\
V DMD/DMD2 & Same mixture as Stage IV & Stage IV checkpoint; Stage I real-score teacher & Asymmetric distribution matching with motion preservation \\
\end{longtable}
}

\subsection{Stage I: Bidirectional Control Adaptation}\label{stage-i}

Stage I keeps full temporal attention and adds the control interface to the pretrained video prior with the flow objective of Eq.~\eqref{eq:flow}. It consists of four consecutive steps, each initialized from the previous checkpoint.

\textbf{I-a, camera.} Starting from Wan2.2-TI2V-5B, LoRA and PRoPE are trained jointly on Control2V. The main run reaches checkpoint 2000 and then performs 400 direction-enhancement updates on \texttt{Control2V\_\allowbreak{}backdown}, giving camera-2400.

\textbf{I-b, action.} From camera-2400, the action branch is trained for 2,000 updates on DROID, where each sample provides \texttt{viewmats}, \texttt{Ks}, 64-D actions, and an embodiment ID. Camera extrinsics in this data are nearly static, so PRoPE is frozen to prevent identity trajectories from becoming a spurious geometric rule.

\textbf{I-c, camera recovery.} Adding the action branch weakens the camera response. Two hundred updates on Control2V with both controls present, training only LoRA, restore it and produce restore-2200. This is the released bidirectional checkpoint; it also serves as the frozen real-score teacher and the fake-score initialization in Stage V.

\textbf{I-d, direction enhancement.} A further 100 updates on \texttt{Control2V\_\allowbreak{}combined}, starting from restore-2200, give \texttt{restore2200\_\allowbreak{}sft000100\_\allowbreak{}full}, which initializes Stage II.

\subsection{Stage II: Block-Causal Conversion by Teacher Forcing}\label{stage-ii}

Stage II is the first stage of the CF/CF++ recipe \cite{ref20,ref21}. The 20-frame latent clip is divided into blocks of four latent frames, and temporal attention becomes block lower-triangular. Each block receives its own timestep: the current block is noised, earlier blocks are provided as clean real history, and later blocks are masked out, so the model learns to predict a block from history alone. Five hundred updates with window 20 and effective batch 8 yield a multistep AR diffusion model. The structural conversion is complete at this point, but the model has only seen clean history (exposure bias) and still needs many sampling steps.

\subsection{Stage III: Online UniPC Trajectory Distillation}\label{stage-iii}

Stage III follows the online causal consistency distillation of Causal Forcing++ \cite{ref21,ref23} and uses \texttt{self\_\allowbreak{}teacher\_\allowbreak{}cd\ +\ online\_\allowbreak{}unipc\_\allowbreak{}distillation}. A frozen copy of the Stage II model is the teacher. Teacher and student start from the same initial noise: the teacher runs a dense 25-step UniPC trajectory, and the student follows the actual 12-step UniPC inference path. For each sample, one student transition is selected at random and kept in the autograd graph, while earlier transitions run under \texttt{no\_\allowbreak{}grad} to preserve the multistep UniPC history. The target is the teacher state nearest the next student noise level. The loss combines Smooth L1 state transfer and framewise temporal-delta Smooth L1 with weight 1.0. Although a shared trainer field retains \texttt{ema\_\allowbreak{}weight=0.99}, this loss explicitly ignores \texttt{ema\_\allowbreak{}model}; the target is the frozen teacher trajectory, not an EMA target.

This stage uses learning rate $2\times10^{-6}$, window 7, sink 1, 20 updates, and effective batch 2. Its output is the causal initialization for Stage IV, which enlarges the operating context to window 20 and sink 4.

\subsection{Stage IV: Mixed-Domain Causal SFT}\label{stage-iv}

Few-step distillation on camera-only data can produce static subjects because the target distribution lacks subject dynamics. Stage IV therefore fine-tunes the causal model on a mixture of Control2V, CrossFPS, DrivingDojo, and PhysicalAI, which adds game control, ego driving, and robot contact. It starts from the Stage III checkpoint with checkpoint overlays for the control modules. The retained continuation spans labels 960 to 1500, or 540 updates, with effective batch 64, window 20, and sink 4. The first latent frame is kept as a clean reference with probability 0.7, so the model is trained in both I2V and T2V modes (Section~\ref{t2v-i2v-generation}). Rolling forcing \cite{ref9} is applied with probability 0.8 in the final segment, exposing the model to its own rolled-out history. Trunk, camera, and action learning rates are $2\times10^{-6}$, $1\times10^{-6}$, and $2\times10^{-5}$.

Input-action conditioning and action-sequence output are independently configurable. The open action-output post-training path can mask input actions and optimize \texttt{{[}F,64{]}} predictions with MSE weight 0.5; video DMD updates the visual generator without requiring this auxiliary objective. The output of this stage is SFT1500.

\subsection{Stage V: Asymmetric DMD/DMD2 with Motion Preservation}\label{dmddmd2-and-motion-preservation}

Stage V uses the DMD score-difference gradient \cite{ref24} and the DMD2 pattern of separately optimizing the generator and a fake-score network that is updated online from current samples \cite{ref11}. The matching is asymmetric: the generator is the causal SFT1500 model, while the real-score model is the frozen bidirectional restore-2200 checkpoint from Stage I, which also initializes the fake-score network. Let \(x_g=G_\theta(\epsilon,c)\). After re-noising at level \(t\), the frozen bidirectional real-score model and trainable fake-score model estimate clean latents \(\hat{x}_0^r\) and \(\hat{x}_0^f\). The implementation uses the per-sample normalized gradient

\[
g_{\mathrm{DMD}}=
\frac{\hat{x}_0^f-\hat{x}_0^r}
{\operatorname{mean}|x_g-\hat{x}_0^r|+\varepsilon}.
\]

Rather than backpropagating through either score network, the generator uses the stop-gradient surrogate

\[
\mathcal{L}_{\mathrm{DMD}}=
\frac{1}{2}\left\|x_g-\operatorname{sg}\!\left(x_g-g_{\mathrm{DMD}}\right)\right\|_2^2.
\]

The fake-score network is optimized separately on re-noised generated samples. Under the flow parameterization its target is \(\epsilon'-x_g\):

\[
\mathcal{L}_{\mathrm{fake}}=
\|v_f(x_t,t,c)-(\epsilon'-x_g)\|_2^2.
\]

To prevent distribution matching from collapsing toward a low-motion conditional mean, we compare mean adjacent-frame latent change in generated and real clips:

\[
\mathcal{L}_{\mathrm{motion}}=
\left|\operatorname{mean}|x_g^i-x_g^{i-1}|-
\operatorname{sg}\!\left(\operatorname{mean}|x_{\mathrm{data}}^i-x_{\mathrm{data}}^{i-1}|\right)\right|,
\quad
\mathcal{L}_G=\mathcal{L}_{\mathrm{DMD}}+\lambda_m\mathcal{L}_{\mathrm{motion}}.
\]

Rollouts use training nodes \texttt{{[}1000,750,500,250{]}} and randomly contain one to five four-frame blocks; only the last block keeps gradients. Real-score guidance is 3.0 and fake-score guidance is zero. Context uses window 20, sink 4, and rolling-forcing probability 0.5. From 1500 to 1515, the generator learning rate is $2\times10^{-6}$ and \(\lambda_m=1\). From 1515 to 1545, they become $1\times10^{-6}$ and 2. The fake-score learning rate remains $1\times10^{-6}$. These four nodes match the four-step generation targeted by distillation; with the current training budget, four-step output remains usable but less sharp and stable, so released inference defaults to eight-step UniPC. Checkpoint 1545 is the released causal model.

\section{Inference and Evaluation}\label{inference-and-evaluation}

\subsection{Default Inference}\label{default-inference}

The released causal model uses eight-step UniPC with CFG 3.0 and generates eight latent frames per block. Four-step sampling is also supported for faster generation at lower quality. Each block attends to a 20-frame local window plus four sink frames. Output is $832\times480$ at 24 fps.

Inference runs in real time on \textbf{one NVIDIA L20 48 GB}: the causal model streams block by block with cross-block KV caching and few-step UniPC sampling, so frames leave the sampler as the rollout is produced rather than after the whole clip is denoised.

\subsection{WBench}\label{wbench}

WBench contains 289 cases, 1,058 interaction rounds, five high-level dimensions, and 22 submetrics \cite{ref17}. Each case is generated as one complete sequence rather than restarted as I2V after every interaction. The benchmark-provided c2w camera is converted to model w2c, intrinsics are normalized, and instructions become a framewise 64-D action stream. Outputs vary from 93 to 861 frames (3.875--35.875 seconds); a four-round case contains 381 frames. Standard resolution is $832\times480$ at 24 fps. Retrieval is enabled only for revisit trajectories. The VLM judge is \texttt{doubao-\allowbreak{}seed-\allowbreak{}2-\allowbreak{}0-\allowbreak{}lite-\allowbreak{}260215} through Volcengine Ark.

\subsection{VBench 1.0}\label{vbench-1.0}

VBench \cite{ref18} is scored with the official VBench 1.0 evaluation code. We generate the official text-to-video suite and the image-to-video suite of VBench++ \cite{ref37} in their official prompt order, without sampling or selecting videos. At the time of writing, 240 of 6,220 T2V videos and 118 of 5,590 I2V videos had been generated and scored (125 frames, $832\times480$, 24 fps, eight-step UniPC). Scores are computed on these videos.

\section{Results}\label{results}

\subsection{WBench Five-Dimensional Scores}\label{wbench-five-dimensional-scores}

Table~\ref{tab:wbench-main} reports the five WBench dimensions and their average. Navi 158 measures navigation only; Full 289 also includes event editing, subject action, and perspective switching, which lowers Interaction.

{\def\LTcaptype{table}
\begin{longtable}[]{@{}
  >{\raggedright\arraybackslash}p{(\linewidth - 12\tabcolsep) * \real{0.1111}}
  >{\raggedleft\arraybackslash}p{(\linewidth - 12\tabcolsep) * \real{0.1481}}
  >{\raggedleft\arraybackslash}p{(\linewidth - 12\tabcolsep) * \real{0.1481}}
  >{\raggedleft\arraybackslash}p{(\linewidth - 12\tabcolsep) * \real{0.1481}}
  >{\raggedleft\arraybackslash}p{(\linewidth - 12\tabcolsep) * \real{0.1481}}
  >{\raggedleft\arraybackslash}p{(\linewidth - 12\tabcolsep) * \real{0.1481}}
  >{\raggedleft\arraybackslash}p{(\linewidth - 12\tabcolsep) * \real{0.1481}}@{}}
\caption{WBench five-dimensional scores of the causal release.}\label{tab:wbench-main}\\
\toprule\noalign{}
\begin{minipage}[b]{\linewidth}\raggedright
Split
\end{minipage} & \begin{minipage}[b]{\linewidth}\raggedleft
Average
\end{minipage} & \begin{minipage}[b]{\linewidth}\raggedleft
Quality
\end{minipage} & \begin{minipage}[b]{\linewidth}\raggedleft
Setting
\end{minipage} & \begin{minipage}[b]{\linewidth}\raggedleft
Interaction
\end{minipage} & \begin{minipage}[b]{\linewidth}\raggedleft
Consistency
\end{minipage} & \begin{minipage}[b]{\linewidth}\raggedleft
Physical
\end{minipage} \\
\midrule\noalign{}
\endfirsthead
\toprule\noalign{}
\begin{minipage}[b]{\linewidth}\raggedright
Split
\end{minipage} & \begin{minipage}[b]{\linewidth}\raggedleft
Average
\end{minipage} & \begin{minipage}[b]{\linewidth}\raggedleft
Quality
\end{minipage} & \begin{minipage}[b]{\linewidth}\raggedleft
Setting
\end{minipage} & \begin{minipage}[b]{\linewidth}\raggedleft
Interaction
\end{minipage} & \begin{minipage}[b]{\linewidth}\raggedleft
Consistency
\end{minipage} & \begin{minipage}[b]{\linewidth}\raggedleft
Physical
\end{minipage} \\
\midrule\noalign{}
\endhead
\bottomrule\noalign{}
\endlastfoot
Navi 158 & 73.5 & 78.2 & 73.5 & 63.4 & 83.6 & 68.6 \\
Full 289 & 70.0 & 78.3 & 73.8 & 47.6 & 82.4 & 68.1 \\
\end{longtable}
}

\subsection{Comparison at Model Scale and Training Compute}\label{comparison-with-wbench-models}

The main advantage of Astronex-World 1.0 is efficiency: a 5B model whose entire post-training runs on two NVIDIA L20 48 GB GPUs. Table~\ref{tab:wbench-compare} therefore compares it with open-source models on the WBench live leaderboard \cite{ref17} (snapshot of September 13, 2026) together with their parameter counts and, where their reports disclose it, their training hardware. Peer scores come from the leaderboard; Astronex-World is evaluated by us with the official code and default VLM judge.

\begin{table}[H]
\centering
\small
\caption{Open-source models on WBench Full 289 with model size and training hardware. Training hardware follows each model's report (n/r: not reported). $^\dagger$: post-trained from a Wan prior.}\label{tab:wbench-compare}
\setlength{\tabcolsep}{3.5pt}
\begin{tabularx}{\linewidth}{@{}Xrlrrrrrr@{}}
\toprule
Model & Params & Training GPUs & Avg. & Qual. & Set. & Inter. & Cons. & Phys. \\
\midrule
Kairos 3.0 \cite{ref64} & 4B & n/r & 65.7 & 73.1 & 70.3 & 41.6 & 83.2 & 60.5 \\
YUME 1.5$^\dagger$ \cite{ref62} & 5B & NVIDIA A100 & 68.9 & 77.6 & 72.4 & 48.4 & 80.9 & 65.4 \\
\textbf{Astronex-World 1.0$^\dagger$} & \textbf{5B} & \textbf{2$\times$ L20 48 GB} & \textbf{70.0} & \textbf{78.3} & 73.8 & 47.6 & 82.4 & 68.1 \\
HY-Video 1.5 \cite{ref61} & 8.3B & n/r & 74.3 & 76.6 & 85.6 & 54.7 & 87.5 & 67.1 \\
LongCat-Video \cite{ref31} & 13.6B & n/r & 69.9 & 77.2 & 72.3 & 45.1 & 86.6 & 68.4 \\
Helios (distilled)$^\dagger$ \cite{ref32} & 14B & 64--128$\times$ H100 & 69.7 & 73.3 & 75.3 & 41.6 & 82.2 & 76.1 \\
LTX-2.3 \cite{ref63} & 22B & n/r & 70.9 & 77.0 & 85.2 & 49.4 & 78.0 & 65.1 \\
\bottomrule
\end{tabularx}
\end{table}

\textbf{Same 5B class.} YUME 1.5 is post-trained from the same Wan2.2 5B prior on NVIDIA A100 GPUs; Astronex-World 1.0, trained on two L20 GPUs, is higher in Average (70.0 against 68.9) and in four of the five dimensions. It is also 4.3 points above Kairos 3.0 (4B).

\textbf{Larger models.} With a third or less of the parameters, Astronex-World 1.0 is above LongCat-Video (13.6B, 69.9) and Helios (14B, 69.7), and within 0.9 points of LTX-2.3 (22B, 70.9). Helios, also post-trained from a Wan prior, uses 64 to 128 H100 GPUs per training stage, while Astronex-World 1.0 reaches a higher Average on two L20 GPUs. Its Quality score of 78.3 is the highest in the table.

\subsection{Capability Comparison with Contemporary World Models}\label{capability-comparison}

Benchmark scores measure the behavior of one release under one protocol; they do not by themselves separate the design choices that distinguish interactive world models. Table~\ref{tab:capability} therefore compares the released capability profile of Astronex-World with five contemporary interactive video world models: LingBot-World \cite{ref33}, HY-World 1.5 (WorldPlay) \cite{ref34}, Genie 3 \cite{ref35}, LongCat-Video \cite{ref31}, and Helios \cite{ref32}. Each entry quotes the corresponding project's own public report, and the figures are not normalized across rows: resolution, frame rate and latency were measured on different hardware, with different step budgets and different definitions of a controllable action.

{\small
\setlength{\tabcolsep}{3pt}
\def\LTcaptype{table} 
\begin{longtable}[]{@{}
  >{\raggedright\arraybackslash}p{(\linewidth - 8\tabcolsep) * \real{0.1400}}
  >{\raggedright\arraybackslash}p{(\linewidth - 8\tabcolsep) * \real{0.2000}}
  >{\raggedright\arraybackslash}p{(\linewidth - 8\tabcolsep) * \real{0.2200}}
  >{\raggedright\arraybackslash}p{(\linewidth - 8\tabcolsep) * \real{0.2200}}
  >{\raggedright\arraybackslash}p{(\linewidth - 8\tabcolsep) * \real{0.2200}}@{}}
\caption{Released capability profile of interactive video world models. Entries follow each project's public report and repository as of September 2026 and are quoted in that project's own terms.}\label{tab:capability}\\
\toprule\noalign{}
Model & Model form & Control interface & Streaming and real-time behavior & Output and long-horizon memory \\
\midrule\noalign{}
\endfirsthead
\toprule\noalign{}
Model & Model form & Control interface & Streaming and real-time behavior & Output and long-horizon memory \\
\midrule\noalign{}
\endhead
\bottomrule\noalign{}
\endlastfoot
\textbf{Astronex-World} & 5B video DiT; bidirectional + causal & Text, image, event, action, camera (PRoPE) & Real-time: 24~fps &$832\times480$ and $1280\times704$; evaluated at 480p \\
LingBot-World \cite{ref33} & Open-source simulator; base + fast & Keyboard actions + camera rotation & Real-time: 16~fps, sub-second latency & 480P / 720P; minute-level horizon \\
HY-World 1.5 (WorldPlay) \cite{ref34} & Streaming video diffusion & Keyboard + mouse & Real-time: 24~fps & 720p; context memory \\
Genie 3 \cite{ref35} & General-purpose world model & Navigation input; text-prompted world events & Real-time: 24~fps & 720p; consistent for minutes \\
LongCat-Video \cite{ref31} & 13.6B DiT; T2V, I2V, continuation & Text, image, video & Not real-time: 720p clip in minutes & 720p, 30~fps; minutes-long clips \\
Helios \cite{ref32} & 14B AR diffusion; T2V, I2V, V2V & Text, image, video & Real-time: 19.5~fps on 1$\times$H100 & Minute-scale; resolution not stated \\
\end{longtable}
}

Three differences follow from Table~\ref{tab:capability}. First, control is expressed at a different level of abstraction. LingBot-World, HY-World 1.5, and Genie 3 take navigation input from a keyboard with camera rotation or mouse, Genie 3 additionally accepts text-prompted world events, and LongCat-Video and Helios are prompt-driven, while Astronex-World carries a 64-dimensional continuous action vector together with an embodiment identifier into every layer and keeps a head that can be post-trained to decode an action sequence from predicted visual states. That interface lets one trunk be reused across control spaces rather than tied to a single input device.

Second, geometry and memory are handled explicitly. PRoPE injects camera intrinsics and extrinsics into all 30 layers, the adapter documents the c2w, w2c and normalized-intrinsics conventions that WBench requires, and long-horizon behavior is carried by a 20-frame window with four persistent sink frames and an optional frustum-overlap retrieval path. The comparison models address the same problem with reconstituted context memory, memory-aware distillation or extended visual memory; Astronex-World combines an explicit projective geometric condition with a window and sink configuration.

Third, the mechanisms behind real-time interaction differ. LingBot-World and HY-World 1.5 distill few-step samplers, Helios compresses the historical and noisy context without KV caching, and Astronex-World combines block-causal KV caching with few-step UniPC sampling and an eight-frame inference block, so the same weights stream block by block on a single L20. The model supports $832\times480$ and $1280\times704$ output; all benchmark results in this report use $832\times480$. Section~\ref{limitations} records the remaining interaction and drift limitations as objectives of the 2.0 release.

\subsection{VBench Results}\label{vbench-results}

\begin{table}[H]
\centering
\small
\caption{VBench 1.0 scores on the partial official runs (240 T2V and 118 I2V videos).}\label{tab:vbench-scores}
\begin{tabular}{@{}lrr@{}}
\toprule
Dimension & Text-to-video & Image-to-video \\
\midrule
Imaging Quality & 0.715 & 0.738 \\
Aesthetic Quality & 0.508 & 0.473 \\
Motion Smoothness & 0.990 & -- \\
Temporal Flickering & 0.987 & 0.995 \\
Dynamic Degree & 0.254 & -- \\
Overall Consistency & 0.220 & -- \\
Background Consistency & -- & 0.985 \\
I2V Background & -- & 0.997 \\
Camera Motion & -- & 0.485 \\
\bottomrule
\end{tabular}
\end{table}

Smoothness, flickering, and background preservation are high, while the low Dynamic Degree reflects the motion suppression after DMD (Section~\ref{internal-motion-diagnostic}).

\subsection{Qualitative Results}\label{qualitative-results}

\begin{figure}[H]
\centering
\includegraphics[width=0.98\linewidth]{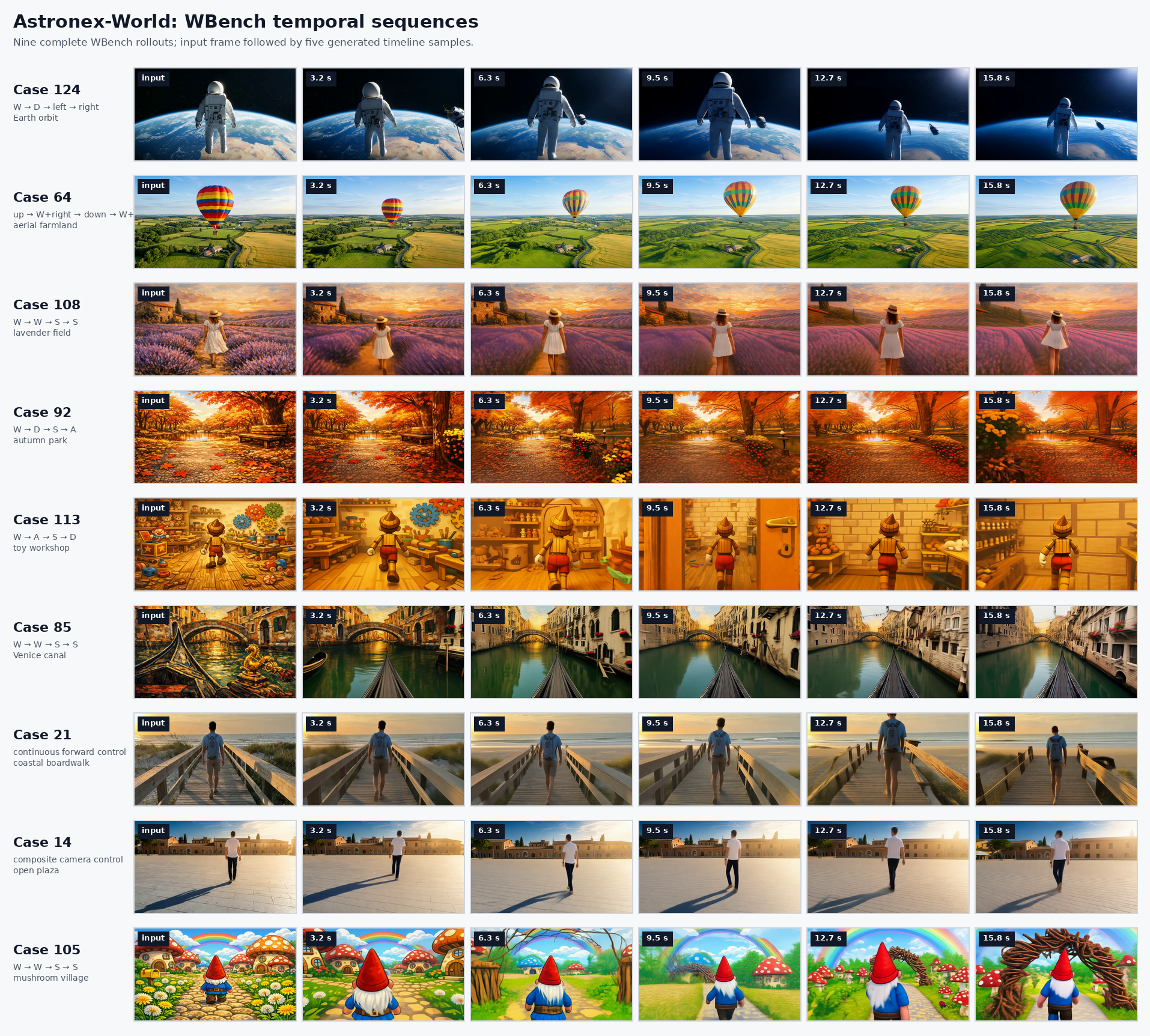}
\caption{WBench temporal sequences. Each row is one complete causal rollout of a four-round case, 381 frames or 15.875 seconds. The first column is the WBench input; five later frames are sampled in temporal order. Nine cases cover orbital Earth, aerial farmland, a lavender field, an autumn park, a toy workshop interior, Venetian canals, a coastal boardwalk, an open plaza, and a mushroom village under forward, backward, lateral, vertical, and composite camera trajectories. The model preserves identity, structure, and dominant color under substantial viewpoint changes; the 158/289 quantitative results provide the aggregate evaluation.}
\label{fig:project-5}
\end{figure}

\begin{figure}[H]
\centering
\includegraphics[width=0.98\linewidth]{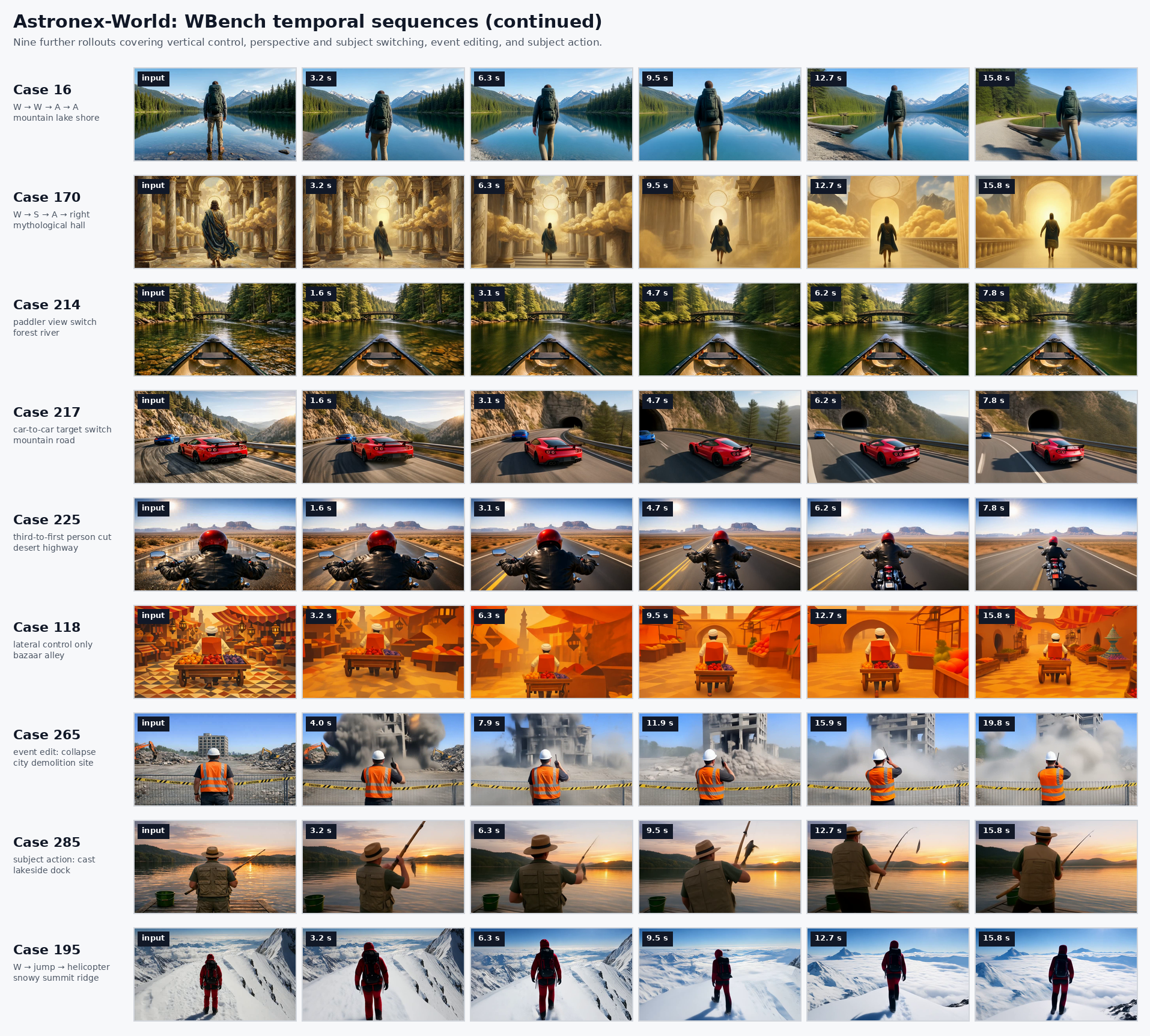}
\caption{Additional WBench temporal sequences, presented in the same format as Figure~\ref{fig:project-5}. Nine further rollouts cover vertical control along a mountain lake shore, a mythological hall, first-person view switching between two paddlers on a forest river, car-to-car target switching on a mountain road, a third-to-first person cut on a desert highway, lateral-only control through a bazaar alley, a staged demolition event chain, a fishing subject action, and a jump followed by an incoming helicopter on a snowy summit ridge. Row length follows the interaction rounds of each case: 381 frames (15.875 seconds) for four-round cases, 189 frames (7.875 seconds) for the two-round view-switch cases, and 477 frames (19.875 seconds) for the event-edit case. Together, Figures~\ref{fig:project-5} and \ref{fig:project-6} show 18 independent WBench rollouts, matching the 18 VBench videos in Figures~\ref{fig:project-9} and \ref{fig:project-10}.}
\label{fig:project-6}
\end{figure}

\begin{figure}[H]
\centering
\includegraphics[width=0.98\linewidth]{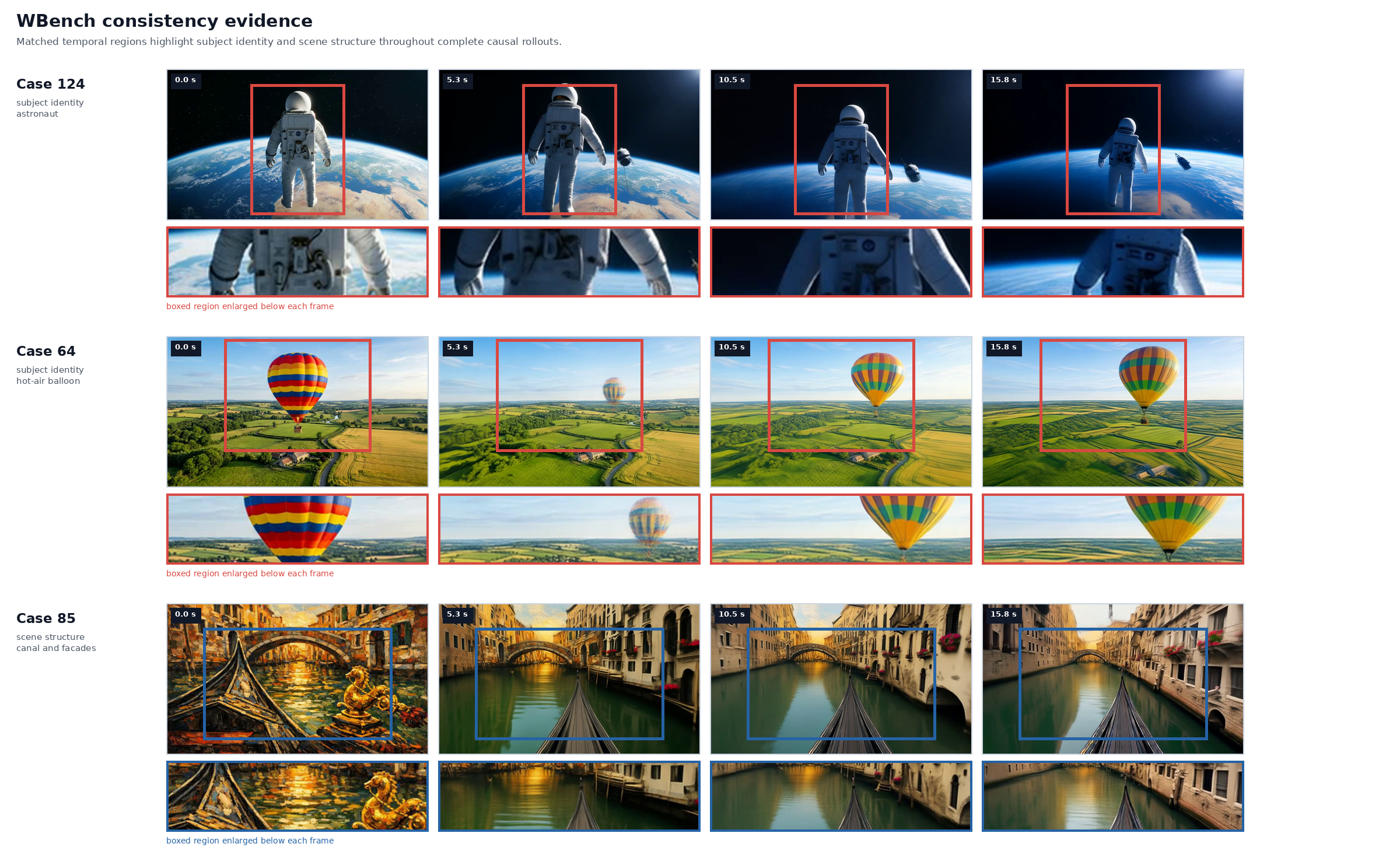}
\caption{Detailed WBench consistency evidence. Red boxes track the principal subject in the astronaut and balloon sequences; blue boxes track canal geometry and facade appearance. The corresponding enlarged regions below each frame expose identity, texture, color, and structural retention across the complete causal rollout.}
\label{fig:project-7}
\end{figure}

\begin{figure}[H]
\centering
\includegraphics[width=0.98\linewidth]{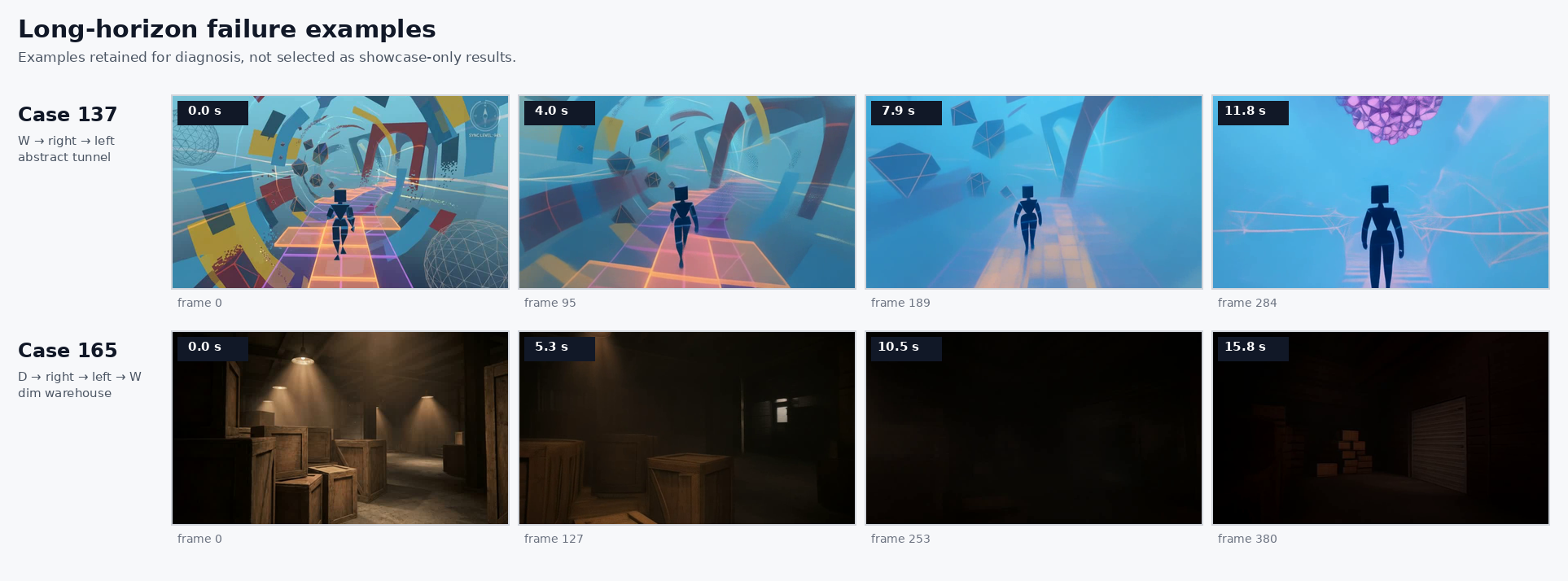}
\caption{Representative failures. In Case 137, the abstract tunnel undergoes color and geometric drift under \texttt{W-right-left}. In Case 165, the warehouse gradually darkens and loses local detail. Attention sinks delay but do not eliminate long-term color and structural drift, and a return turn does not guarantee reversible scene evolution.}
\label{fig:project-8}
\end{figure}

\begin{figure}[H]
\centering
\includegraphics[width=0.98\linewidth]{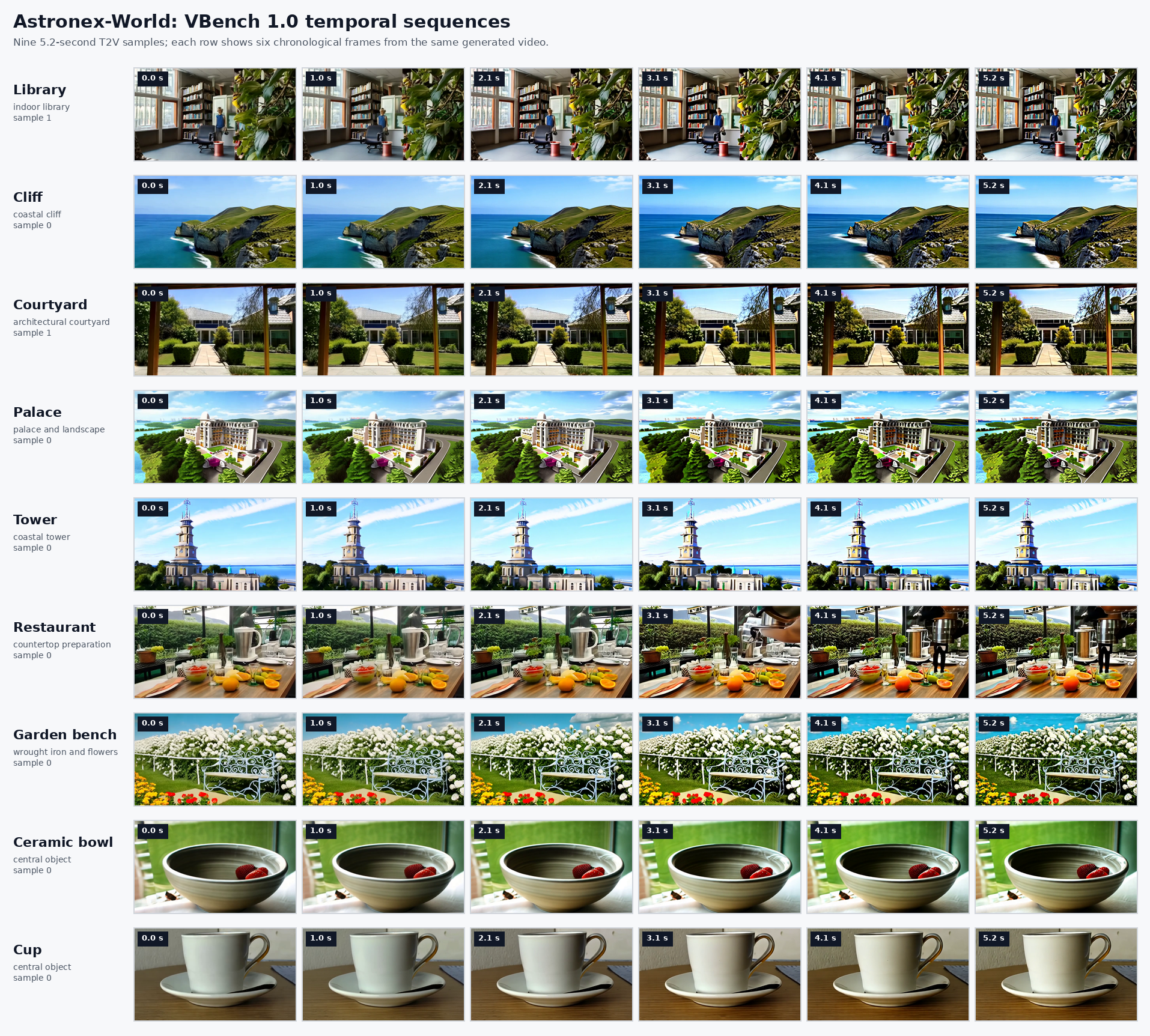}
\caption{VBench 1.0 T2V sequences. Each row is an independent 125-frame, 5.208-second text-only video, with six frames shown chronologically and no input image. The first nine examples cover a library, sea cliff, courtyard, palace, coastal tower, restaurant counter, garden bench, ceramic bowl, and cup and demonstrate text-conditioned scene diversity, object persistence, and temporal continuity.}
\label{fig:project-9}
\end{figure}

\begin{figure}[H]
\centering
\includegraphics[width=0.98\linewidth]{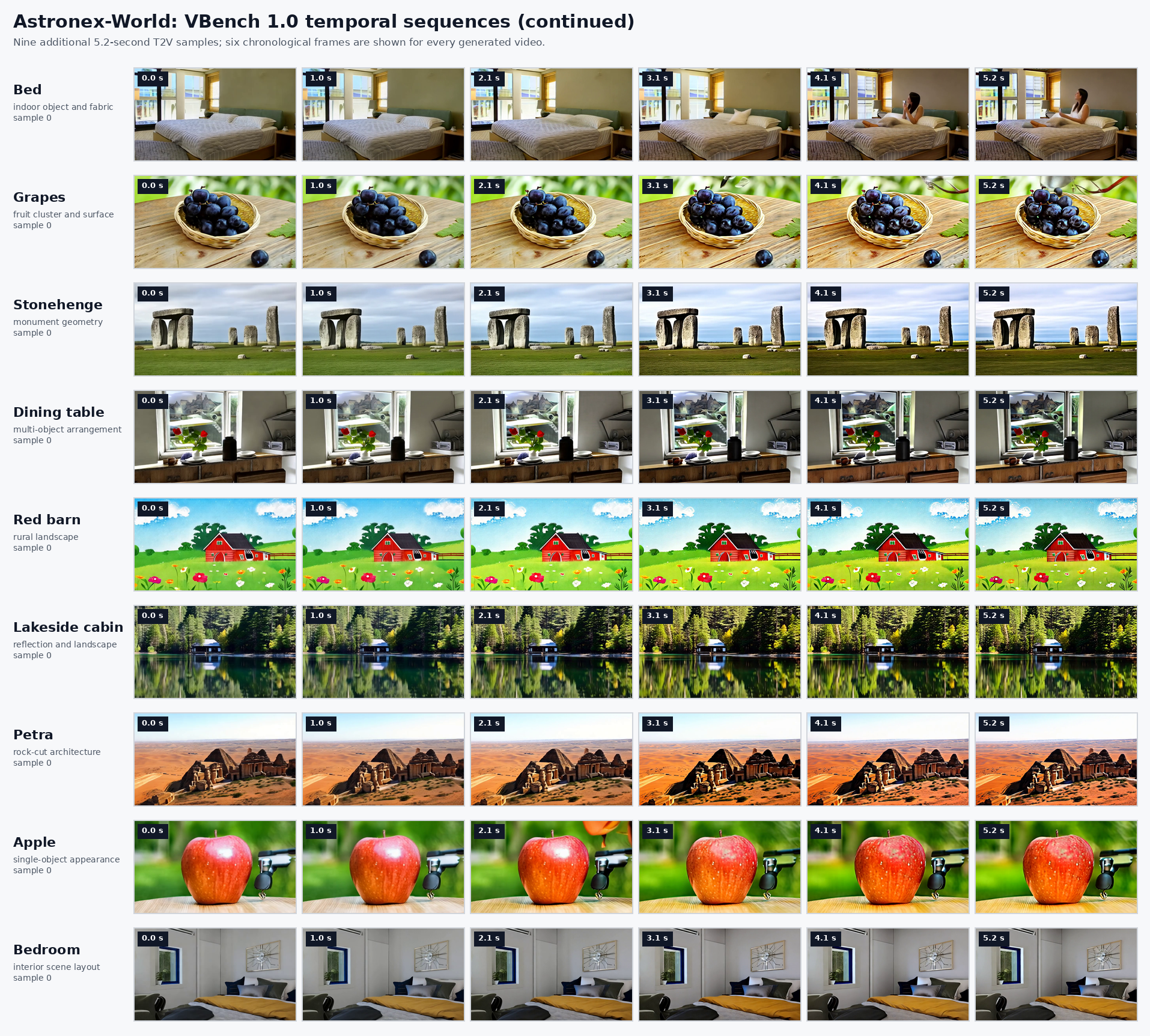}
\caption{Additional VBench 1.0 T2V sequences. Nine further videos cover a bed, grapes, Stonehenge, dining table, red barn, lakeside cabin, Petra, apple, and bedroom. Together, Figures~\ref{fig:project-9} and \ref{fig:project-10} show 18 independent VBench videos spanning single objects, multi-object arrangements, interiors, architecture, monuments, and outdoor landscapes.}
\label{fig:project-10}
\end{figure}

\begin{figure}[H]
\centering
\includegraphics[width=0.98\linewidth]{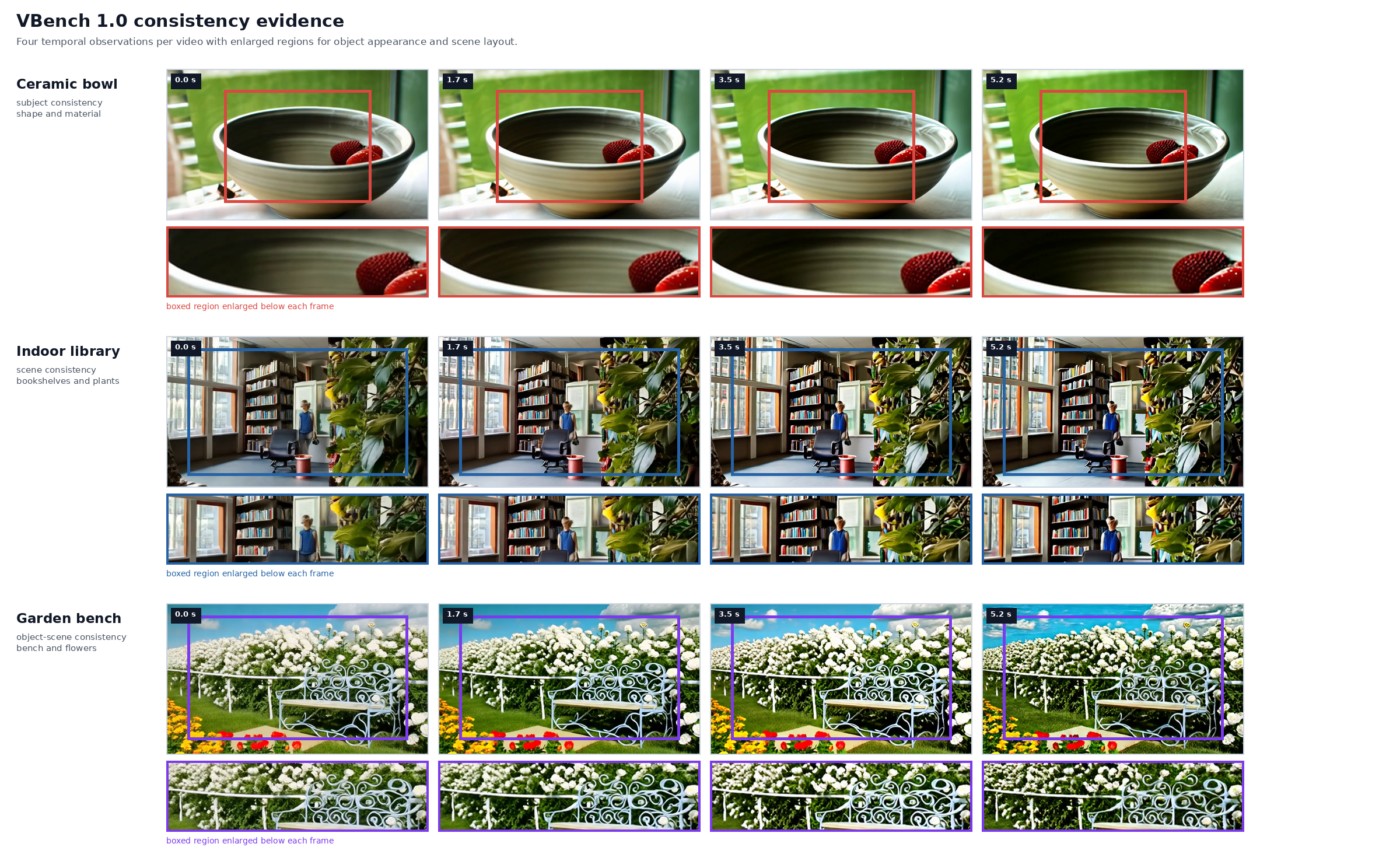}
\caption{Detailed VBench consistency evidence. Red boxes enlarge the ceramic bowl to compare shape and material across time; blue boxes expose the persistent library layout; purple boxes jointly track the garden bench and surrounding flowers. Every column comes from a different time in the same generated video.}
\label{fig:project-11}
\end{figure}

Figures~\ref{fig:project-5} and \ref{fig:project-6} present 18 complete WBench causal rollouts, and Figures~\ref{fig:project-9} and \ref{fig:project-10} present 18 VBench videos, so both benchmarks contribute the same number of qualitative examples. Figures~\ref{fig:project-7}, \ref{fig:project-8}, and \ref{fig:project-11} add matched consistency details and failure diagnostics.

\subsection{Analysis}\label{analysis}

Visual quality and consistency are the clearest strengths: Quality 78.3 and Consistency 82.4 on Full 289, supported by block-causal KV caching, a local window, and persistent sinks. Interaction is the main weakness. It is 63.4 on Navi 158 but 47.6 on Full 289, because event editing, subject action, and perspective switching are much harder for the current text-switch interface. Physical 68.1 leaves room for more accurate contact and object dynamics.

\subsection{Downstream Adaptation}\label{downstream-adaptation}

The bidirectional and causal models form a general visual-dynamics foundation. Its text, image, camera, action, and action-output interfaces allow downstream work to reuse the trunk while adapting data and control mappings. For autonomous driving, ego trajectory, steering, speed, and throttle/brake can be written into the 64-D action stream, and event prompts can insert traffic events at chosen times. For embodied intelligence, robot trajectories can post-train the action-output head as an action-sequence decoder \cite{ref25}, with adapters mapping the 64-D interface to joints, end effectors, grippers, or mobile bases. Both directions reuse the released weights and interfaces directly.

\subsection{Internal Motion Diagnostic}\label{internal-motion-diagnostic}

A single basketball sequence is used during development to measure subject residual amplitude. This is not a public benchmark and is not comparable across models, but it reveals DMD over-staticization.

{\def\LTcaptype{table}
\begin{longtable}[]{@{}lrr@{}}
\caption{Subject residual amplitude on the internal basketball sequence.}\label{tab:motion}\\
\toprule\noalign{}
Model state & Residual amplitude & Relative to bidirectional teacher \\
\midrule\noalign{}
\endfirsthead
\toprule\noalign{}
Model state & Residual amplitude & Relative to bidirectional teacher \\
\midrule\noalign{}
\endhead
\bottomrule\noalign{}
\endlastfoot
Bidirectional & 88.8 px & 100\% \\
SFT1500 & 36.1 px & 41\% \\
DMD1515 & 21.2 px & 24\% \\
DMD1545 & 16.6 px & 19\% \\
\end{longtable}
}

Final DMD improves few-step quality and overall consistency but strongly suppresses subject motion. The motion term prevents complete freezing without restoring bidirectional-teacher amplitude, consistent with the moderate Visual Plausibility and Subject Action scores.

\section{Limitations}\label{limitations}

\begin{enumerate}
\def\labelenumi{\arabic{enumi}.}
\tightlist
\item
  \textbf{Weak complex interaction.} Training emphasizes continuous camera and action streams. Event editing, subject action, and perspective switching rely on a limited text-switch interface that cannot express multiple independent timed events.
\item
  \textbf{Remaining long-horizon drift.} Four sink frames and local history delay forgetting, but strong turns, loops, and low-light scenes still exhibit color shift, darkening, structural repainting, and detail loss.
\item
  \textbf{Distillation suppresses motion.} DMD may prefer stable but static solutions; internal diagnostics remain far below bidirectional motion amplitude.
\item
  \textbf{Limited physical fidelity.} The model has no explicit physical state, depth, collision constraints, or 3D scene representation. Video coherence is not proof of accurate physics.
\item
  \textbf{Action post-training.} The 64-D action stream and \texttt{{[}F,64{]}} action-output path are reserved interfaces; driving and robot control spaces are connected through post-training with domain data.
\end{enumerate}

\section{Future Work: Astronex-World 2.0}\label{future-work-astronex-world-2.0}

Astronex-World 2.0 will include 14B and 5B scales. The 14B model targets greater visual capacity, complex dynamics, and long-horizon scene modeling; the 5B model retains the low-deployment-cost and real-time-oriented path. Both scales will use unified text, image, camera, continuous-action, and embodiment interfaces and will retain bidirectional and causal forms.

A primary objective is to address long-term color drift, structural repainting, loop inconsistency, and motion decay. Future training will combine full-sequence bidirectional representations with causal self-rollout distributions, longer sequences, cross-block consistency supervision, geometric and photometric constraints, and systematic memory training. ``Combination'' here means a unified model family, control interface, and objectives, not parallel execution of two models at inference.

Building on the reserved action interface, post-trained models for embodied intelligence and autonomous driving will follow, with action adherence, planning, and task success added to the evaluation.

\section{Conclusion}\label{conclusion}

Astronex-World 1.0 is a 5B controllable video world-model foundation comprising bidirectional and causal few-step models. The bidirectional model retains full temporal modeling; the causal model supports persistent generation through block-causal attention, KV caching, online trajectory distillation, and DMD/DMD2 distribution matching. Both forms share text, image, camera, action, and event interfaces.

\textbf{Performance at the 5B scale.} The causal model scores 73.5 on WBench Navi 158 and 70.0 on Full 289, with Quality 78.3, Setting 73.8, Interaction 47.6, Consistency 82.4, and Physical 68.1. Among 5B-class open models, it is above YUME 1.5, which is post-trained from the same Wan2.2 5B prior (70.0 against 68.9, and higher in four of the five dimensions), and above the 4B Kairos 3.0 by 4.3 points. Against larger open models, it is above LongCat-Video (13.6B) and Helios (14B) and within 0.9 points of LTX-2.3 (22B), and its Quality score is the highest among the compared models.

\textbf{Results on two L20 GPUs.} All five training stages run on two NVIDIA L20 48 GB GPUs, and the released causal model streams $832\times480$ video at 24 fps in real time on a single L20. With this budget, Astronex-World 1.0 obtains bidirectional and causal releases with camera, action, and event control, covers all 289 WBench cases, and exceeds Helios, which uses 64 to 128 H100 GPUs per training stage with a 14B model.

\textbf{What this shows.} A strong pretrained video prior combined with a staged post-training recipe---control adaptation, block-causal conversion, trajectory distillation, mixed-domain SFT, and asymmetric DMD---is enough to build a real-time interactive world model that competes with models two to four times larger. Large GPU clusters and larger parameter counts are therefore not a prerequisite for this level of interactive world modeling, and the recipe is reproducible on hardware available to most research groups.

The reserved action input and output interfaces provide a post-training starting point for embodied intelligence and autonomous driving. Astronex-World 2.0 will extend 14B and 5B models toward complex events, subject actions, perspective switching, long-term color stability, and motion preservation. This 1.0 report provides a reproducible model, training, and evaluation baseline while keeping the interfaces open for subsequent research.


\addcontentsline{toc}{section}{References}
\begin{thebibliography}{99}
\bibitem{ref1} Wan Team. \href{https://arxiv.org/abs/2503.20314}{Wan: Open and Advanced Large-Scale Video Generative Models}. arXiv:2503.20314, 2025. Wan2.2 implementation: \href{https://github.com/Wan-Video/Wan2.2}{Wan-Video/Wan2.2}.
\bibitem{ref6} Ruilong Li et al. \href{https://arxiv.org/abs/2507.10496}{Cameras as Relative Positional Encoding}. NeurIPS, 2025.
\bibitem{ref43} David Ha, J\"urgen Schmidhuber. \href{https://arxiv.org/abs/1803.10122}{World Models}. arXiv:1803.10122, 2018.
\bibitem{ref50} Anthony Hu et al. \href{https://arxiv.org/abs/2309.17080}{GAIA-1: A Generative World Model for Autonomous Driving}. arXiv:2309.17080, 2023.
\bibitem{ref52} Shenyuan Gao et al. \href{https://arxiv.org/abs/2405.17398}{Vista: A Generalizable Driving World Model with High Fidelity and Versatile Controllability}. NeurIPS, 2024.
\bibitem{ref53} NVIDIA. \href{https://arxiv.org/abs/2501.03575}{Cosmos World Foundation Model Platform for Physical AI}. arXiv:2501.03575, 2025.
\bibitem{ref54} Sherry Yang et al. \href{https://arxiv.org/abs/2310.06114}{Learning Interactive Real-World Simulators}. ICLR, 2024.
\bibitem{ref55} Yilun Du et al. \href{https://arxiv.org/abs/2302.00111}{Learning Universal Policies via Text-Guided Video Generation}. NeurIPS, 2023.
\bibitem{ref44} Danijar Hafner et al. \href{https://arxiv.org/abs/2301.04104}{Mastering Diverse Domains through World Models}. arXiv:2301.04104, 2023.
\bibitem{ref45} Jake Bruce et al. \href{https://arxiv.org/abs/2402.15391}{Genie: Generative Interactive Environments}. ICML, 2024.
\bibitem{ref46} Dani Valevski et al. \href{https://arxiv.org/abs/2408.14837}{Diffusion Models Are Real-Time Game Engines}. ICLR, 2025.
\bibitem{ref47} Eloi Alonso et al. \href{https://arxiv.org/abs/2405.12399}{Diffusion for World Modeling: Visual Details Matter in Atari}. NeurIPS, 2024.
\bibitem{ref48} Xianglong He et al. \href{https://arxiv.org/abs/2508.13009}{Matrix-Game 2.0: An Open-Source Real-Time and Streaming Interactive World Model}. arXiv:2508.13009, 2025.
\bibitem{ref49} Xiaofeng Mao et al. \href{https://arxiv.org/abs/2507.17744}{Yume: An Interactive World Generation Model}. arXiv:2507.17744, 2025.
\bibitem{ref34} Tencent Hunyuan Team. \href{https://arxiv.org/abs/2512.14614}{WorldPlay: Towards Long-Term Geometric Consistency for Real-Time Interactive World Modeling}. arXiv:2512.14614, 2025. Project page: \href{https://3d-models.hunyuan.tencent.com/world/}{3d-models.hunyuan.tencent.com/world}.
\bibitem{ref33} Robbyant Team. \href{https://arxiv.org/abs/2601.20540}{Advancing Open-source World Models} (LingBot-World). arXiv:2601.20540, 2026. Code and models: \href{https://github.com/Robbyant/lingbot-world}{Robbyant/lingbot-world}.
\bibitem{ref35} Google DeepMind. \href{https://deepmind.google/discover/blog/genie-3-a-new-frontier-for-world-models/}{Genie 3: A New Frontier for World Models}. Blog post, 2025.
\bibitem{ref51} Lloyd Russell et al. \href{https://arxiv.org/abs/2503.20523}{GAIA-2: A Controllable Multi-View Generative World Model for Autonomous Driving}. arXiv:2503.20523, 2025.
\bibitem{ref56} Mido Assran et al. \href{https://arxiv.org/abs/2506.09985}{V-JEPA 2: Self-Supervised Video Models Enable Understanding, Prediction and Planning}. arXiv:2506.09985, 2025.
\bibitem{ref2} Jonathan Ho, Ajay Jain, Pieter Abbeel. \href{https://arxiv.org/abs/2006.11239}{Denoising Diffusion Probabilistic Models}. NeurIPS, 2020.
\bibitem{ref3} William Peebles, Saining Xie. \href{https://arxiv.org/abs/2212.09748}{Scalable Diffusion Models with Transformers}. ICCV, 2023.
\bibitem{ref5} Edward J. Hu et al. \href{https://arxiv.org/abs/2106.09685}{LoRA: Low-Rank Adaptation of Large Language Models}. ICLR, 2022.
\bibitem{ref4} Yaron Lipman et al. \href{https://arxiv.org/abs/2210.02747}{Flow Matching for Generative Modeling}. ICLR, 2023.
\bibitem{ref39} Xingchao Liu, Chengyue Gong, Qiang Liu. \href{https://arxiv.org/abs/2209.03003}{Flow Straight and Fast: Learning to Generate and Transfer Data with Rectified Flow}. ICLR, 2023.
\bibitem{ref38} Patrick Esser et al. \href{https://arxiv.org/abs/2403.03206}{Scaling Rectified Flow Transformers for High-Resolution Image Synthesis}. ICML, 2024.
\bibitem{ref28} Ashish Vaswani et al. \href{https://arxiv.org/abs/1706.03762}{Attention Is All You Need}. NeurIPS, 2017.
\bibitem{ref7} Tianwei Yin et al. \href{https://arxiv.org/abs/2412.07772}{From Slow Bidirectional to Fast Autoregressive Video Diffusion Models}. arXiv:2412.07772, 2024.
\bibitem{ref8} Xun Huang et al. \href{https://arxiv.org/abs/2506.08009}{Self Forcing: Bridging the Train-Test Gap in Autoregressive Video Diffusion}. arXiv:2506.08009, 2025.
\bibitem{ref22} Boyuan Chen et al. \href{https://arxiv.org/abs/2407.01392}{Diffusion Forcing: Next-token Prediction Meets Full-Sequence Diffusion}. NeurIPS, 2024.
\bibitem{ref59} Sand AI. \href{https://arxiv.org/abs/2505.13211}{MAGI-1: Autoregressive Video Generation at Scale}. arXiv:2505.13211, 2025.
\bibitem{ref60} Guibin Chen et al. \href{https://arxiv.org/abs/2504.13074}{SkyReels-V2: Infinite-length Film Generative Model}. arXiv:2504.13074, 2025.
\bibitem{ref20} Hongzhou Zhu et al. \href{https://arxiv.org/abs/2602.02214}{Causal Forcing: Autoregressive Diffusion Distillation Done Right for High-Quality Real-Time Interactive Video Generation}. ICML, 2026.
\bibitem{ref21} Min Zhao et al. \href{https://arxiv.org/abs/2605.15141}{Causal Forcing++: Scalable Few-Step Autoregressive Diffusion Distillation for Real-Time Interactive Video Generation}. Technical Report, 2026.
\bibitem{ref23} Yang Song et al. \href{https://arxiv.org/abs/2303.01469}{Consistency Models}. ICML, 2023.
\bibitem{ref27} Guangxuan Xiao et al. \href{https://arxiv.org/abs/2309.17453}{Efficient Streaming Language Models with Attention Sinks}. ICLR, 2024.
\bibitem{ref9} Kunhao Liu et al. \href{https://arxiv.org/abs/2509.25161}{Rolling Forcing: Autoregressive Long Video Diffusion in Real Time}. arXiv:2509.25161, 2025.
\bibitem{ref10} Shuai Yang et al. \href{https://arxiv.org/abs/2509.22622}{LongLive: Real-time Interactive Long Video Generation}. ICLR, 2026.
\bibitem{ref40} Jiwen Yu et al. \href{https://arxiv.org/abs/2506.03141}{Context as Memory: Scene-Consistent Interactive Long Video Generation with Memory Retrieval}. arXiv:2506.03141, 2025.
\bibitem{ref41} Zeqi Xiao et al. \href{https://arxiv.org/abs/2504.12369}{WorldMem: Long-term Consistent World Simulation with Memory}. arXiv:2504.12369, 2025.
\bibitem{ref42} Runjia Li et al. \href{https://arxiv.org/abs/2506.18903}{VMem: Consistent Interactive Video Scene Generation with Surfel-Indexed View Memory}. arXiv:2506.18903, 2025.
\bibitem{ref57} Hao He et al. \href{https://arxiv.org/abs/2404.02101}{CameraCtrl: Enabling Camera Control for Text-to-Video Generation}. arXiv:2404.02101, 2024.
\bibitem{ref58} Zhouxia Wang et al. \href{https://arxiv.org/abs/2312.03641}{MotionCtrl: A Unified and Flexible Motion Controller for Video Generation}. arXiv:2312.03641, 2023.
\bibitem{ref24} Tianwei Yin et al. \href{https://arxiv.org/abs/2311.18828}{One-step Diffusion with Distribution Matching Distillation}. CVPR, 2024.
\bibitem{ref11} Tianwei Yin et al. \href{https://arxiv.org/abs/2405.14867}{Improved Distribution Matching Distillation for Fast Image Synthesis}. NeurIPS, 2024.
\bibitem{ref12} Yuanzhi Zhu et al. \href{https://arxiv.org/abs/2412.05899}{Accelerating Video Diffusion Models via Distribution Matching}. arXiv:2412.05899, 2024.
\bibitem{ref13} Wenliang Zhao et al. \href{https://arxiv.org/abs/2302.04867}{UniPC: A Unified Predictor-Corrector Framework for Fast Sampling of Diffusion Models}. NeurIPS, 2023.
\bibitem{ref19} Min Zhao et al. \href{https://arxiv.org/abs/2605.30263}{minWM: A Full-Stack Open-Source Framework for Real-Time Interactive Video World Models}. Technical Report, 2026.
\bibitem{ref30} MIN-Lab. \href{https://huggingface.co/datasets/Junyi42/minwm-data}{minWM-data}. Hugging Face Dataset, 2026.
\bibitem{ref26} Alexander Khazatsky et al. \href{https://arxiv.org/abs/2403.12945}{DROID: A Large-Scale In-The-Wild Robot Manipulation Dataset}. Robotics: Science and Systems, 2024.
\bibitem{ref14} Zizhao Tong et al. \href{https://arxiv.org/abs/2605.23345}{SCOPE: Simulating Cross-game Operations in Playable Environments for FPS World Models}. arXiv:2605.23345, 2026.
\bibitem{ref15} Yuqi Wang et al. \href{https://arxiv.org/abs/2410.10738}{DrivingDojo Dataset: Advancing Interactive and Knowledge-Enriched Driving World Model}. NeurIPS, 2024.
\bibitem{ref16} NVIDIA. \href{https://huggingface.co/datasets/nvidia/PhysicalAI-WorldModel-Synthetic-Embodied-Robot-Scenes}{PhysicalAI WorldModel Synthetic Embodied Robot Scenes}. Dataset, 2026.
\bibitem{ref36} Jonathan Ho, Tim Salimans. \href{https://arxiv.org/abs/2207.12598}{Classifier-Free Diffusion Guidance}. arXiv:2207.12598, 2022.
\bibitem{ref25} Ruixiang Wang et al. \href{https://arxiv.org/abs/2603.17808}{EVA: Aligning Video World Models with Executable Robot Actions via Inverse Dynamics Rewards}. arXiv:2603.17808, 2026.
\bibitem{ref17} Kaining Ying et al. \href{https://arxiv.org/abs/2605.25874}{WBench: A Comprehensive Multi-turn Benchmark for Interactive Video World Model Evaluation}. arXiv:2605.25874, 2026. Code: \href{https://github.com/meituan-longcat/WBench}{meituan-longcat/WBench}.
\bibitem{ref18} Ziqi Huang et al. \href{https://arxiv.org/abs/2311.17982}{VBench: Comprehensive Benchmark Suite for Video Generative Models}. CVPR, 2024.
\bibitem{ref37} Ziqi Huang et al. \href{https://arxiv.org/abs/2411.13503}{VBench++: Comprehensive and Versatile Benchmark Suite for Video Generative Models}. arXiv:2411.13503, 2024.
\bibitem{ref31} Meituan LongCat Team. \href{https://arxiv.org/abs/2510.22200}{LongCat-Video Technical Report}. arXiv:2510.22200, 2025.
\bibitem{ref32} Shenghai Yuan et al. \href{https://arxiv.org/abs/2603.04379}{Helios: Real Real-Time Long Video Generation Model}. arXiv:2603.04379, 2026.
\bibitem{ref61} Tencent Hunyuan Foundation Model Team. \href{https://arxiv.org/abs/2511.18870}{HunyuanVideo 1.5 Technical Report}. arXiv:2511.18870, 2025.
\bibitem{ref62} Xiaofeng Mao et al. \href{https://arxiv.org/abs/2512.22096}{Yume1.5: A Text-Controlled Interactive World Generation Model}. arXiv:2512.22096, 2025.
\bibitem{ref63} Lightricks. \href{https://arxiv.org/abs/2601.03233}{LTX-2: Efficient Joint Audio-Visual Foundation Model}. arXiv:2601.03233, 2026. LTX-2.3 weights: \href{https://huggingface.co/Lightricks/LTX-2.3}{Lightricks/LTX-2.3}.
\bibitem{ref64} Kairos Team. \href{https://arxiv.org/abs/2606.16533}{Kairos: A Native World Model Stack for Physical AI}. arXiv:2606.16533, 2026. Code: \href{https://github.com/kairos-agi/kairos-sensenova}{kairos-agi/kairos-sensenova}.
\end{thebibliography}
\end{document}